%% file: main.tex
\documentclass[10pt,twocolumn,letterpaper]{article}

\usepackage{iccv}              

\usepackage{algorithm}
\usepackage{algorithmicx}
\usepackage{algpseudocode}
 

\definecolor{iccvblue}{rgb}{0.21,0.49,0.74}
\usepackage[pagebackref,breaklinks,colorlinks,allcolors=iccvblue]{hyperref}

\def\paperID{14696} 
\def\confName{ICCV}
\def\confYear{2025}

\title{Bi-VLM: Pushing Ultra-Low Precision Post-Training Quantization Boundaries in Vision-Language Models}

\author{Xijun Wang$^*$, Junyun Huang$^*$, Rayyan Abdalla$^*$, Chengyuan Zhang, Ruiqi Xian, Dinesh Manocha\\
University of Maryland, College Park, USA\\
{\tt\small xijun@umd.edu}
}

\begin{document}
\maketitle
\input{sec/0_abstract}    
\input{sec/1_intro}
\input{sec/2_related}
\input{sec/3_method}
\input{sec/4_experiment}
\input{sec/5_conclusion}

{
\small
\bibliographystyle{ieeenat_fullname}
\bibliography{main}
}

\input{sec/6_appendix}
\end{document}

%% file: sec/0_abstract.tex
\begin{abstract}

We address the critical gap between the computational demands of vision-language models and the possible ultra-low-bit weight precision (bitwidth $\leq2$ bits) we can use for higher efficiency.  Our work is motivated by the substantial computational cost and memory requirements of VLMs, which restrict their applicability in hardware-constrained environments.  We propose Bi-VLM, which separates model weights non-uniformly based on the Gaussian quantiles. Our formulation groups the model weights into outlier (salient) and multiple inlier (unsalient) subsets, ensuring that each subset contains a proportion of weights corresponding to its quantile in the distribution. We propose a saliency-aware hybrid quantization algorithm and use it to quantize weights by imposing different constraints on the scaler and binary matrices based on the saliency metric and compression objective.  We have evaluated our approach on different VLMs. For the language model part of the VLM, our Bi-VLM outperforms the SOTA by 3\%-47\% on the visual question answering task in terms of four different benchmarks and three different models. For the overall VLM,  our Bi-VLM outperforms the SOTA by 4\%-45\%. We also perform token pruning on the quantized models and observe that there is redundancy of image tokens 90\% - 99\% in the quantized models. This helps us to further prune the visual tokens to improve efficiency.

{\renewcommand\thefootnote{}\footnotetext{*Equal Contribution.}}

\end{abstract}

%% file: sec/1_intro.tex
\section{Introduction}
\label{sec:intro}

In recent years, vision-language models (VLMs) have emerged as powerful tools for performing multi-modal tasks by combining visual and textual information. The pretrained VLMs, LLaVA~\cite{liu2024visual} and Qwen~\cite{Qwen2.5-VL}, have demonstrated remarkable performance on a wide range of benchmarks, including single-image analysis, multi-image interpretation, and video understanding. These achievements are achieved by their vast model parameters and extensive training data, enabling high accuracy and generalization across diverse tasks. However, they have high computational demands and high memory usage, which pose significant challenges for deployment on resource-constrained devices such as wearables, mobile devices, and FPGAs~\cite{zeng2024flightllm}.



Post-training quantization (PTQ)~\cite{nagel2019data,nagel2020up,krishnamoorthi2018quantizing} has gained significant traction for VLMs due to its efficiency and practicality. Unlike quantization-aware training (QAT)~\cite{Jacob2017QuantizationAT,gupta2015deep}, which requires access to the training datasets, PTQ operates on frozen network parameters and uses a small calibration set to determine optimal rounding functions. Recently, PTQ methods~\cite{frantar2022gptq, lin2024awq} have achieved promising results by reducing the bitwidths for weights and activations. 

Despite notable advances in 8-bit and 4-bit quantization for VLMs~\cite{dettmers2023spqr, xiao2023smoothquant, wang2024q}, the increasing size and complexity of these models require more aggressive quantization strategies. Neural network binarization, which reduces the bit width of the weight to a single bit, is a promising direction to achieve ultra-low bit quantization~\cite{helwegen2019latent, qin2020forward, qin2023bibench}. However, existing PTQ methods may not work in terms of ultra-low-bit quantization ($\leq$2 bits), leading to substantial performance degradation. State-of-the-art binary PTQ approaches, such as PB-LLM~\cite{shang2023pb}, provide limited performance levels with significant trade-offs in accuracy. We need better ultra-low-bit quantization methods for VLMs that can preserve the task performance. 

{\bf Main Results:} We present a novel approach that uses ultra-low-bit precision for VLMs. We first evaluate the sensitivity of different components in VLMs, including vision encoder, adaptor/projector, and language model. As shown in Tables~\ref{table:components_llama},~\ref{table:components_llava}, and ~\ref{table:components_qwen}, the vision model exhibits high sensitivity to quantization (on average); the adaptor/projector exhibits less sensitivity to quantization, barely affecting the performance; the language model exhibits considerable sensitivity to quantization. 
Our empirical analysis, presented in Supplementary \ref{appendix:A}, reveals that most weight values exhibit a near-Gaussian distribution. Furthermore, outlier density varies across layers, necessitating an adaptive quantization approach that assigns higher precision to critical weights while aggressively binarizing most of the model to maximize compression efficiency. Additionally, outliers, constitute about 5\% of Vision Model weights and 1\% of Language Model weights, as shown in Supplementary \ref{appendix:B}, and have a substantial impact on the model's performance. As a result, any uniform quantization strategy may not work well. 

Based on this analysis, we design an efficient quantization strategy for VLMs that analyzes key weight distribution properties, including saliency, sparsity, and outliers. 
In particular, we present a saliency-aware hybrid quantization algorithm, where salient weights receive higher bit precision and unsalient weights are binarized, ensuring a balance between storage efficiency and quantization error minimization.
Our approach, Bi-VLM, separates model weights non-uniformly based on Gaussian quantiles. Model weights are grouped into outlier (salient) and multiple inlier (unsalient) subsets, ensuring each subset contains a proportion of weights corresponding to its respective quantile in the distribution. We use our saliency-aware hybrid quantization algorithm to quantize the weights, where we solve the quantization problem by imposing different constraints on the scaler and binary matrices based on the saliency metric and compression objective. 

To the best of our knowledge, ours may the first work to explore the ultra-low-bit post training quantization ($\leq$2 bits) for VLMs. We also present strategies for achieving aggressive bit-width reduction while maintaining robust performance across diverse benchmarks. Some of the key contributions of our work include:

\begin{enumerate}
    \item We evaluate the sensitivity of different components in VLMs. On average, the language model exhibits considerable sensitivity to quantization.  
    \item We propose Bi-VLM which separates weights non-uniformly based on Gaussian quantiles and then use our proposed saliency-aware hybrid quantization to quantize the weights.
    \item We conduct token pruning on the quantized models, and observe that there is 90\%-99\% image token redundancy even in the quantized models. This helps us further prune the visual tokens to improve the efficiency.
    \item We push post-training quantization to bit-level for large VLMs in terms of four different benchmarks and three different model series. From our experiments, our Bi-VLM outperforms the SOTA by 3\%-47\% on the language model part of the VLM and 4\%-45\% for the overall VLMs.

\end{enumerate}

%% file: sec/2_related.tex
\section{Related Works}

\subsection{Post-Quantization on VLM}
Post-training quantization (PTQ) has become a practical solution for efficiently deploying Vision-Language Models (VLMs), as it replaces full-precision tensors with low-precision values, significantly reducing storage requirements and computational overhead. Unlike quantization-aware training (QAT), which demands costly retraining and access to the training dataset, PTQ methods leverage small calibration sets to optimize rounding functions with reduced data requirements and computational costs.


GPTQ~\cite{frantar2022gptq} employs Hessian-based second-order error compensation to minimize block-wise quantization errors, achieving excellent performance at ultra-low bit-widths (e.g., 4-bit quantization). Techniques such as AWQ~\cite{lin2024awq} and OWQ~\cite{Lee2023OWQOW} further enhance PTQ by preserving critical weight channels and scaling them appropriately for activation features, thereby retaining information representation capacity. Additionally, methods like PB-LLM\cite{shang2023pb} and SpQR~\cite{dettmers2023spqr} adopt feature segmentation strategies, selectively quantizing non-critical components to mitigate performance loss while maintaining low bit-width. QLoRA~\cite{dettmers2024qlora} introduces a memory-efficient fine-tuning method by backpropagating through frozen 4-bit quantized models using Low Rank Adapters (LoRA) and proposes innovations like 4-bit NormalFloat (NF4), double quantization, and paged optimizers to minimize memory usage without sacrificing performance. Q-VLM~\cite{wang2024q} proposes a post-training quantization approach for large vision-language models (LVLMs) that optimizes cross-layer dependency using activation entropy as a proxy, enabling block-wise partitioning to reduce discretization errors and search costs while maintaining performance efficiency.

While these methods demonstrate success in efficiently quantizing VLMs, they often rely on layer-wise or block-wise optimization, which overlooks dependencies across layers or components of the model. Addressing these cross-layer interactions remains a key challenge for achieving optimal quantization performance in large-scale VLMs.

\subsection{Network Binarization}

Binarization reduces neural network parameters to a single bit, significantly decreasing storage requirements and computational costs. This is achieved by converting full-precision weights into binary values using the sign function, combined with a scaling factor to minimize binarization errors. Typically, the scaling factor is applied in a channel-wise manner to better preserve information~\cite{rastegari2016xnor, qin2023bibench}.

Most existing binarization methods rely on quantization-aware training (QAT), where the training process accounts for quantization effects. Straight Through Estimator (STE)~\cite{bengio2013estimating} is often used to overcome gradient vanishing issues caused by the non-differentiable nature of the binarization function. Binary Weight Network (BWN)~\cite{rastegari2016xnor} was among the first to demonstrate binarized weights while maintaining full-precision activations, whereas XNOR-Net~\cite{rastegari2016xnor} extended this approach by binarizing both weights and activations for higher efficiency. Methods like DoReFa-Net~\cite{zhou2016dorefa} further improved the training speed by introducing quantized gradients.

Recent advancements have explored group-wise binarization strategies, where network weights are divided into smaller groups to minimize binarization errors~\cite{faraone2018syq}. Notably, binarization techniques have been successfully applied to Transformers~\cite{wang2023bitnet} and BERT models~\cite{qin2022bibert}, demonstrating the feasibility of deploying binarized networks in real-world scenarios.

For large language models (LLMs), PB-LLM~\cite{shang2023pb} investigates the use of binarized QAT and post-training quantization (PTQ) strategies. However, it reveals the challenge of retaining a significant portion of weights, typically over 30\%, at INT8 precision to maintain acceptable performance. BiLLM~\cite{huang2024billm} aims to push the boundaries of PTQ-based binarization for LLMs. By minimizing reliance on higher bit-widths while ensuring performance, these methods make significant strides toward fully binarized LLMs, offering a promising solution for resource-efficient deployment in large-scale applications.

%% file: sec/3_method.tex
\section{Our Approach}

\subsection{Binarization Formulation}  

Network binarization is a quantization technique that restricts weight values to discrete binary levels, typically \(\{-1,1\}\). The purpose is to reduce memory footprint and computational complexity while maintaining accuracy. To enhance expressiveness, it is often combined with low-bit quantization, where more sensitive weights are assigned higher precision.  

We propose a generalized formulation of the weight quantization process that unifies both binarization and low-bit quantization as a discrete optimization problem, aiming to find the best binary representation of weights while introducing a row-wise scaling factor to preserve reconstruction accuracy. Given a weight matrix \( W \in \mathbb{R}^{m \times n} \), we minimize the reconstruction error:

\begin{equation} \label{opt_eq}
\|W - A B\|^2_F,
\end{equation}
where \( A \in \mathbb{R}^{m \times m} \) is a diagonal scaling matrix, defined as \( A = \text{diag}(a_{11}, a_{22}, ..., a_{mm}) \), ensuring row-wise scaling. The binary matrix \( B \in \mathbb{R}^{m \times n} \) contains elements \( b_{ij} \) constrained to take \( N_b \) discrete values within the range \( [-1,1] \). For storage purposes, each \( b_{ij} \) is later mapped to a set of quantized values \( \{0, 1, ..., 2^{N_b} -1\} \), where \( N_b \) represents the number of bits assigned per weight element.  

The objective function in Equation~\eqref{opt_eq} is minimized with respect to the Frobenius norm \( \|\cdot\|_F \), which reduces the element-wise squared differences between \( W \) and \( AB \), providing an optimal approximation by minimizing residual energy. This formulation is particularly effective when outliers are not dominant, as confirmed by empirical analysis.  
For \( N_b = 1 \), the constraints in Equation~\eqref{opt_eq} reduce to \( b_{ij} \in \{-1,1\} \), corresponding to the standard binarization process.

\subsection{Bi-VLM}
\subsubsection{Quantile-Based Weight Partitioning}

\begin{figure}[htb]
    \centering
    \includegraphics[width=0.9\linewidth]{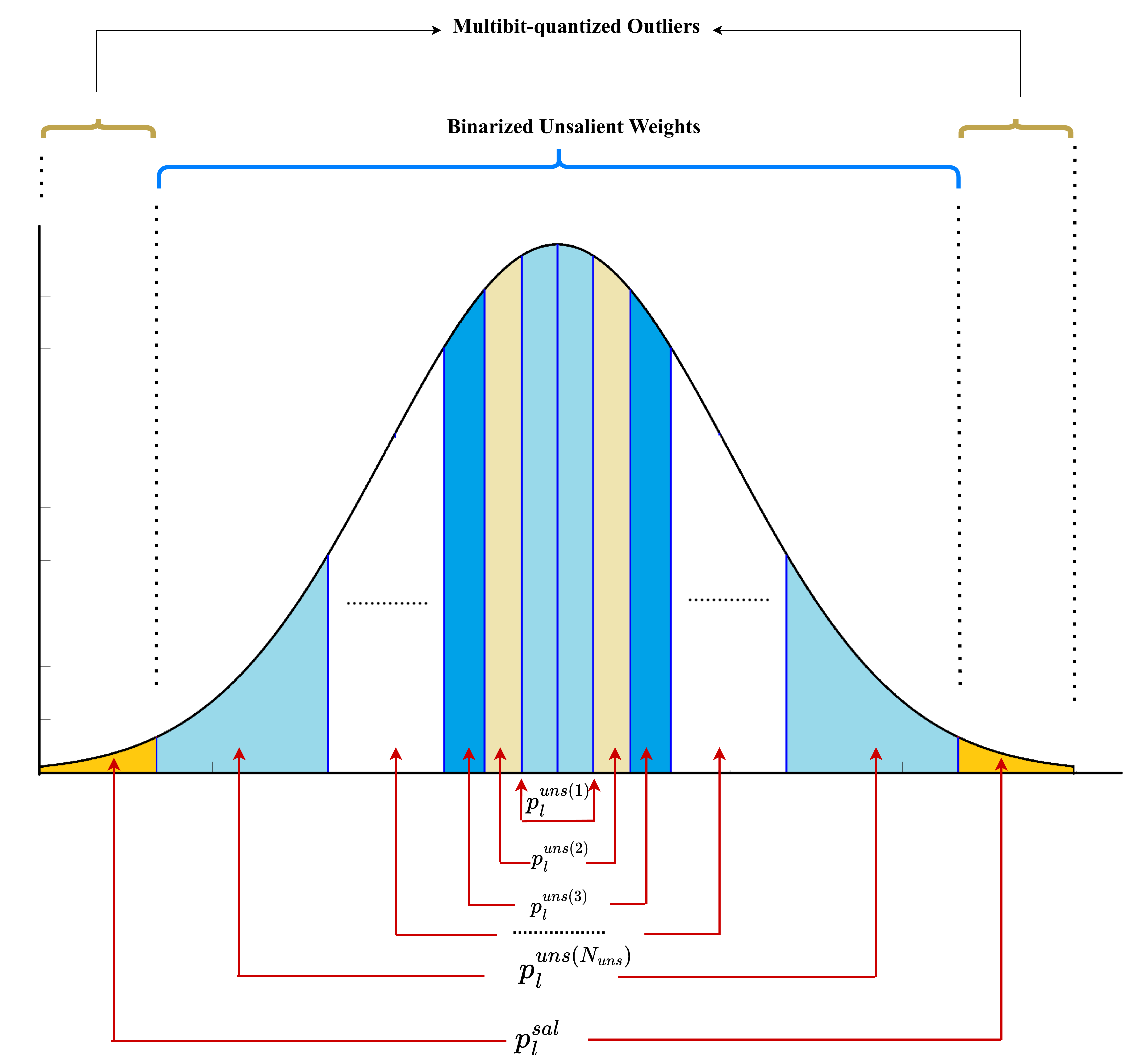}  
    \vspace{-2mm}
    \caption{Saliency-aware quantile-based partitioning of Gaussian-distributed weights. Unsalient weights are divided into equal quantiles and binarized, while salient weights, corresponding to the distribution tails, are quantized using a multi-bit approach.}
    \label{fig:fig_prune}
    \vspace{-2mm}
\end{figure}

A major challenge in quantizing large models, such as LLMs and VLMs, is identifying salient weights—those whose quantization may significantly impact performance. Saliency is typically assessed using either magnitude or the Hessian-based metric, though the latter is computationally expensive with limited benefits \cite{shang2023pb}.

We use a magnitude-based criterion to classify model weights within each layer into distinct subsets. Instead of linear partitioning, we divide the weights non-uniformly based on Gaussian quantiles. Specifically, weights are grouped into outlier (salient) and multiple inlier (unsalient) subsets, ensuring each subset contains a proportion of weights corresponding to its respective quantile in the distribution.

Given a model with \( N_l \) layers, where each layer is indexed by \( l = 1,2,\dots,N_l \), we analyze the weight values of each layer \( l \) and define the set of all weights as \( \mathcal{W}_l \). We partition \( \mathcal{W}_l \) into salient weights \( \mathcal{S}_l \) and unsalient weights \( \mathcal{S}_l^c \), such that:

\begin{equation}
\mathcal{W}_l = \mathcal{S}_l \cup \mathcal{S}_l^c, \quad \mathcal{S}_l \cap \mathcal{S}_l^c = \varnothing.
\end{equation}
Furthermore, we divide the unsalient set \( \mathcal{S}_l^c \) into \( N_{\text{uns}} \) disjoint subsets \( \mathcal{S}_l^{c(k)} \), indexed by \( k = 1, \dots, N_{\text{uns}} \), such that:

\begin{equation}
\mathcal{S}_l^c = \bigcup_{k=1}^{N_{\text{uns}}} \mathcal{S}_l^{c(k)}.
\end{equation}
Our classification, guided by Gaussian quantiles, partitions the weights into \( \mathcal{S}_l^c \) and \( \mathcal{S}_l^{c(k)} \) for \( k = 1, \dots, N_{\text{uns}} \). This partitioning is based on the layer-specific mean \( \mu_l \) and standard deviation \( \sigma_l \) and assumes, based on empirical evidence, that the weight distribution in each layer is symmetric about its mean. Consequently, quantile analysis is performed on the right side of the distribution and mirrored to the left.

We define the percentile \( p^{sal}_{\text{l}} \) as the proportion of the highest-magnitude weights classified as salient. This corresponds to the upper quantile of the Gaussian distribution. Consequently, the total quantile of the unsalient region is \( 1 - p_{\text{l}}^{sal} \). Since this region is divided into \( N_{\text{uns}} \) subsets, each subset corresponds to a quantile of size:

\begin{equation}
p_{l}^{uns} = \frac{1 - p_{l}^{sal}}{N_{\text{uns}}},
\end{equation}
In order to determine the magnitude cutoff values for salient and unsalient subsets, we compute the z-scores from the standard normal distribution corresponding to the respective quantiles. These are given by:

\begin{equation}
z^{(k)} = \Phi^{-1} \left(\frac{1 + k \cdot p_l^{\text{uns}}}{2} \right), \quad k = 1,2,\dots,N_{\text{uns}},
\end{equation}
where \( \Phi^{-1}(\cdot) \) is the \textit{probit function}, i.e., the inverse cumulative distribution function (CDF) of the Gaussian distribution. Therefore, the salient and non-salient subsets for layer \( l \) are defined as:

\begin{equation}
\begin{aligned}
    \mathcal{S}_l &= \{ w \in \mathcal{W}_l \mid |w| > \mu_l + \sigma_l z^{(N_{\text{uns}})} \} \\
    \mathcal{S}_l^{c(k)} &= \{ w \in \mathcal{W}_l \mid \mu_l + \sigma_l z^{(k-1)} < |w| \leq \mu_l + \sigma_l z^{(k)} \},
\end{aligned}
\end{equation}
where \( \mu_l , \sigma_l \) are the mean and standard deviation of the weight distribution in layer \( l \), and \( z^{k} \) is the z-score corresponding to the upper magnitude boundary of the non-salient region indexed \(k = 1,2,\dots,N_{\text{uns}}\). Overall, our approach enables independent analysis of layer weights while parameterizing the proportion of salient weights and the number of non-salient partitions to optimize the accuracy-compression tradeoff.

\subsubsection{Saliency-Aware Hybrid Quantization}

We present a hybrid quantization method applied independently to each model layer. The weight matrix \( W_l \) of layer \( l \) is decomposed into a salient weight matrix \( W_l^{\text{sal}} \in \mathbb{R}^{m \times n} \) and multiple unsalient weight component matrices \( W_l^{\text{uns}(k)} \in \mathbb{R}^{m \times n} \) for \( k = 1,2,\dots,N_{\text{uns}} \), enabling distinct processing strategies for quantization. 
\begin{equation} 
W_l = [w_{ij}]_{1 \leq i \leq m, 1 \leq j \leq n} = W_l^{\text{sal}} + \sum_{k=1}^{N_{\text{uns}}} W_l^{\text{uns}(k)},
\end{equation}
where \( W_l^{\text{sal}} \) and \( W_l^{\text{uns}(k)} \) are defined as:
\begin{equation} \label{wsal}
W_l^{\text{sal}} = \{w^{\text{sal}}_{l,ij} \mid w_{l,ij} = w_{ij} \text{ if } w_{ij} \in \mathcal{S}, \text{ else } 0\},  
\end{equation}
\begin{equation} \label{wuns}
W_l^{\text{uns}(k)} = \{w_{l,ij}^{{\text{uns}}(k)} \mid w_{l,ij}^{(k)} = w_{ij} \text{ if } w_{ij} \in \mathcal{S}_l^{c(k)}, \text{ else } 0\}.
\end{equation}
In order to optimize compression while preserving accuracy, we solve the quantization problem in Equation~(\ref{opt_eq}) separately for the salient and unsalient components, imposing different constraints on the matrices \( A \) and \( B \) based on the saliency metric and compression objective. Specifically, salient weights, which are highly sensitive and impact the model's performance, are quantized using two bits, whereas unsalient weights within the inlier subsets are binarized using a single bit. 

\paragraph{Salient Weights Two-Bit Quantization:} \
The salient weight matrix \( W_l^{\text{sal}} \) (Equation~\ref{wsal}) is quantized into a binary matrix \( B_l^{\text{sal}} \) using \( N_b = 2 \) bits, along with a row-wise scaling vector \( \mathbf{a} \in \mathbb{R}^m \), yielding:

\begin{equation}
W_l^{\text{sal,quantized}} = \mathbf{a} \odot B_l^{\text{sal}}.
\end{equation}
Here, \( \odot \) denotes row-wise multiplication. The optimization problem (Equation~\ref{opt_eq}) is solved with \( B = B_l^{\text{sal}} \), where elements take \(2^{N_{bit}}\) discrete values in \( [-1,1] \) and are mapped to the set \( \{0,1,2,3\} \), binary represented for storage. The scaling matrix is:
\begin{equation}
A_l^{\text{sal}} = \text{diag}(a_1^{\text{sal}}, \dots, a_m^{\text{sal}}).
\end{equation}
Thus, the quantization problem is formulated as:
\begin{equation} \label{sal_opt_eq}
\|W_l^{\text{sal}} - \mathbf{a} \odot B_l^{\text{sal}}\|_F^2.
\end{equation}
We analyze the convexity of Equation~(\ref{sal_opt_eq}) by reformulating it as a row-wise optimization:
\begin{equation}
\|w_{l,ij}^{\text{sal}} - a_i^{\text{sal}} b_{l,ij}^{\text{sal}}\|_2^2, \quad \forall \ 1 \leq j \leq n.
\end{equation}
Since the \( \ell_2 \)-norm is convex, the optimization formulation is convex with respect to the scaling parameter \( a_i^{\text{sal}} \). However, the discrete nature of \( B_l^{\text{sal}} \), constrained to \( 2^{N_b} \) values, results in a discrete optimization problem where gradient-based solvers fail due to the discontinuous solution space. To address this, we relax \( b_{l,ij}^{\text{sal}} \) to the continuous interval \( [-1,1] \), enabling an iterative quadratic programming approach that alternates between optimizing \( a_i^{\text{sal}} \) and updating \( B_l^{\text{sal}} \), ensuring a feasible quantization solution. After convergence, \( B_l^{\text{sal}} \) is mapped back to a discrete set of \( 2^{N_b} \) values, corresponding to the midpoints of \( 2^{N_b} + 1 \) quantization levels, as described in Algorithm \ref{alg:sal}.

Our empirical analysis reveals that outlier weights contribute unevenly to the tails of the Gaussian distribution, necessitating adaptive quantization resolution. To address this, we apply an exponential adaptation to the \( 2^{N_b} + 1 \) linearly spaced levels \( d \in [-1,1] \) after convex optimization:

\begin{equation} \label{adaptive_quant}
r_{\text{levels}}(d) = \mu_{B_l^{\text{sal}}}^{\text{optimal}} + \sigma_{B_l^{\text{sal}}}^{\text{optimal}} \cdot \text{sign}(d) \cdot \left( \alpha \times \exp(|d|) - 1 \right),
\end{equation}
where \( r_{levels} \) is the adaptively defined quantization level corresponding to the linearly defined level \( d \). Quantization centers, \( r_{\text{centers}} \), representing the midpoints between each subsequent level, encode mapped values from \( B_l^{\text{sal}} \) as a binary value of length \( N_b = 2 \). The parameters \( \mu_{B_l^{\text{sal}}}^{\text{optimal}} \) and \( \sigma_{B_l^{\text{sal}}}^{\text{optimal}} \) denote the mean and standard deviation of the optimized \( B_l^{\text{sal}} \), respectively. The term \( \text{sign}(d) \) ensures symmetric quantization. The parameter \( \alpha \), empirically set to 1.4, adjusts the mapping to provide finer resolution for small weights and coarser representation for larger values.

\begin{algorithm}[h] 
\caption{Row-wise Quantization with Scale and Binary Matrix}
\label{alg:sal}
\begin{algorithmic}[1]
\Require Weight matrix \( W \in \mathbb{R}^{m \times n} \), iterations \( \text{iters} \), bits \( N_b \)
\Ensure Scale factors \( a \in \mathbb{R}^{m} \), quantized matrix \( B \in \mathbb{R}^{m \times n} \)

\State \textbf{Initialize:} \( B \gets \text{sign}(W) \), \( a \gets \mathbf{0} \)

\For{\( \text{iter} = 1 \) to \( \text{iters} \)}
    \State \( a_{\text{old}} \gets a \) 
    \State \( a \gets \frac{\sum_{j} W_{ij} B_{ij}}{\sum_{j} B_{ij}^2} \) where \( \sum_{j} B_{ij}^2 > 0 \), else \( a_i = 0 \)
    \State \( B \gets \text{clip} \left( \frac{W}{a[:, None]}, -1, 1 \right) \)

\EndFor

\State \textbf{Adaptive Quantization Step:}
\State Compute \( \mu_B, \sigma_B \) from nonzero elements of \( B \)
\State Compute quantization levels using \eqref{adaptive_quant}
\State \( r_{\text{centers}} \gets \frac{r_{\text{levels}}[:-1] + r_{\text{levels}}[1:]}{2} \)
\State \( B \gets r_{\text{centers}}[\arg\min |B - r_{\text{centers}}|] \)

\State \Return \( a \), \( B \)

\end{algorithmic}
\end{algorithm}

\paragraph{Unsalient Weights Binarization:} \ 
With salient weights quantized and separated, the remaining components of the weight matrix in layer \( l \) correspond to the unsalient weights. We apply strict binarization by mapping each matrix \( W_l^{\text{uns}(k)} \) to a binary matrix \( B_l^{\text{uns}(k)} \) and a scalar factor \( a_l^{\text{uns}(k)} \), such that:
\begin{equation}\label{uns_q}
W_l^{\text{uns,quantized}(k)} = a_l^{\text{uns}(k)}  \times B_l^{\text{uns}(k)}.
\end{equation}
Reformulating the optimization problem in \eqref{opt_eq}, we set \( B = B_l^{\text{uns}(k)} \) and \( A = a_l^{\text{uns}(k)} I_{m \times m} \), where \( B_l^{\text{uns}(k)} \in \{-1,1\} \). The resulting binarization problem can be expressed as:

\begin{equation}\label{uns_opt}
\|W_l^{\text{uns}(k)} - a_l^{\text{uns}(k)} B_l^{\text{uns}(k)} \|_F^2.
\end{equation}
Since \( b_{l,ij}^{\text{uns}(k)} \) is restricted to \( \{-1,1\} \), an element-wise thresholding strategy minimizes the squared error, yielding the optimal binary values:

\begin{equation}
b_{l,ij}^{\text{uns}(k)*} =
\begin{cases} 
\text{sign}(w_{l,ij}^{\text{uns}(k)}) & \text{if } a_l^{\text{uns}(k)} > 0, \\
-\text{sign}(w_{l,ij}^{\text{uns}(k)}) & \text{if } a_l^{\text{uns}(k)} < 0.
\end{cases}
\end{equation}
With \( B_l^{\text{uns}(k)} \) now predetermined, a solution to the convex optimization formulation of the quadratic program in \eqref{sal_opt_eq} with respect to \( a_l^{\text{uns}(k)} \) gives an optimal solution:
\begin{equation} \label{a_uns}
a_{l,ij}^{\text{uns}(k)*} = \frac{\langle W_l^{\text{uns}(k)}, B_l^{\text{uns}(k)} \rangle}{\|B_l^{\text{uns}(k)}\|_F^2},
\end{equation}
where the inner product is \( \langle W_l^{\text{uns}(k)}, B_l^{\text{uns}(k)} \rangle = \text{tr} \left( W_l^{\text{uns}(k)^\top} B_l^{\text{uns}(k)} \right).\) This guarantees that \( a_{l,ij}^{\text{uns}(k)} \) is always positive, ensuring optimal binary matrix:

\begin{equation} \label{Buns}
B_{l,ij}^{\text{uns}(k)*} = \text{sign}(W_{l,ij}^{\text{uns}(k)}).
\end{equation}

\begin{algorithm}[h]
\caption{Unsalient Weights Binarization}
\label{alg:unsalient_binarization}
\begin{algorithmic}[1]
\Require Weight matrix \( W_l^{\text{uns}(k)} \in \mathbb{R}^{m \times n} \)
\Ensure Binary matrix \( B_l^{\text{uns}(k)} \in \{-1,1\}^{m \times n} \) and scaling factor \( a_l^{\text{uns}(k)} \in \mathbb{R} \)

\State \textbf{Step 1: Compute Binary Matrix}  
\State Compute \( B_l^{\text{uns}(k)} \) using Equation~\eqref{Buns}

\State \textbf{Step 2: Compute Scaling Factor}  
\State Compute \( a_l^{\text{uns}(k)} \) using Equation~\eqref{a_uns}

\State \Return \( B_l^{\text{uns}(k)} \), \( a_l^{\text{uns}(k)} \)
\end{algorithmic}
\end{algorithm}

\subsubsection{Adaptive Saliency Search via Optimization}

Salient weights deviate significantly from the typical weight distribution, making them unsuitable for binarization. Their inclusion in the unsalient subset \( \mathcal{S}_l^c \) leads to rectification, degrading model performance \cite{shang2023pb}, while selecting too many salient weights in \( \mathcal{S}_l \) increases storage requirements, reducing compression efficiency.

In order to balance the performance and compression, we formulate salient weight selection as a numerical optimization problem that determines the optimal salient percentile \( p_l^{\text{sal}} \). The optimized saliency region is constrained by a compression threshold set by the maximum proportion of salient weights quantized with \( N_b \) bits and the number of binarized unsalient quantiles \( N_{\text{uns}} \). Given a specific layer \( l \) with weight matrix \( W_l \), a predefined unsalient region partition count \( N_{\text{uns}} \), and a maximum allowable salient percentile \( p_l^{\text{sal}, \max} \), we minimize the following objective function based on normalized reconstruction error:
%
%
\begin{align}
    &\min_{p_l^{\text{sal}}} \mathcal{J}(p_l^{\text{sal}}; W_l, N_{\text{uns}}, N_b) = \\ \notag
    &\frac{\|W_l^{\text{sal}} - \mathbf{a} \odot B_l^{\text{sal}}\|_F^2 + \sum_{k=1}^{N_{\text{uns}}} \|W_l^{\text{uns}(k)} - a_l^{\text{uns}(k)} B_l^{\text{uns}(k)} \|_F^2}{\|W_l\|_F^2}, 
\end{align}
\[
\text{s.t.} \quad p_l^{\text{sal}} \in [0, p_l^{\text{sal}, \max}].
\]
Here, \( W_l^{\text{sal}} \) and \( W_l^{\text{uns}(k)} \) are obtained from Equations~\eqref{wsal} and~\eqref{wuns}, respectively, while the pairs \( \{ {\mathbf{a} ,B_l^{\text{sal}}} \}\) and \( \{ a_l^{\text{uns}(k)} ,B_l^{\text{uns}(k)} \} \) are computed using Algorithms~\ref{alg:sal} and~\ref{alg:unsalient_binarization}.
Since the Frobenius norm \( \|\cdot\|_F^2 \) is convex and our hybrid quantization approach relaxes the discrete space of \( B_l^{\text{sal}} \) and \( B_l^{\text{uns}(k)} \), the objective function \( \mathcal{J} \) ensures a global minimum for given \( W_l \), \( N_{\text{uns}} \), and \( N_b \). For computational efficiency, we employ bounded numerical optimization using Brent’s method, a gradient-free approach that combines golden-section search for robustness and parabolic interpolation for fast convergence \cite{brent2013algorithms}. Once the optimal salient percentile \( p_l^{\text{sal}, \text{opt}} \) is determined, we apply the full quantization process to layer \( W_l \), as outlined in Algorithm~\ref{alg3}.

\begin{algorithm}[h]
\caption{Our Overall Method}
\label{alg3}
\begin{algorithmic}[1]

\Require \( W_l \in \mathbb{R}^{m \times n} \), \( N_{\text{uns}} \), \( N_b \), \( p_l^{\text{sal}, \max} \)
\Ensure \( W_l^{\text{quantized}} \)

\State \textbf{Step 1: Initialize} \( p_l^{\text{sal}} = p_l^{\text{sal}, \max} \)

\State \textbf{Step 2: Optimize} Minimize \( \mathcal{J} \) (Eq.~\eqref{sal_opt_eq}): \textbf{Hybrid Quantization} \( (W_l, N_{\text{uns}}, N_b, p_l^{\text{sal}}) \) over \( p_l^{\text{sal}} \in [0, p_l^{\text{sal}, \max}] \), find \( p_l^{\text{sal, opt}} \)

\State \textbf{Step 3: Quantize} Run \textbf{Hybrid Quantization} \( (W_l, N_{\text{uns}}, N_b, p_l^{\text{sal, opt}}) \)

\State \textbf{Step 4: Reconstruct} 
\[
W_l^{\text{quantized}} = W_l^{\text{sal,quantized}} + \sum_{k=1}^{N_{\text{uns}}} W_l^{\text{uns,quantized}(k)}
\]

\State \Return \( W_l^{\text{quantized}} \)

\Statex

\textbf{Hybrid Quantization} \( (W_l, N_{\text{uns}}, N_b, p_l^{\text{sal}}) \)

\State Partition \( W_l \) into \( W_l^{\text{sal}} \) and \( W_l^{\text{uns}(k)} \) (Eqs.~\eqref{wsal}, \eqref{wuns})
\State Quantize \( W_l^{\text{sal}} \) into \( B_l^{\text{sal}}, \mathbf{a} \) (Algorithm ~\ref{alg:sal})
\For{\( k = 1 \) to \( N_{\text{uns}} \)}
    \State Quantize \( W_l^{\text{uns}(k)} \) into \( B_l^{\text{uns}(k)}, a_l^{\text{uns}(k)} \) (Algorithm ~\ref{alg:unsalient_binarization})
\EndFor
\State \Return \( B_l^{\text{sal}}, \mathbf{a}, \{(a_l^{\text{uns}(k)}, B_l^{\text{uns}(k)})\}_{k=1}^{N_{\text{uns}}} \)

\end{algorithmic}
\end{algorithm}

\input{tables/llama}
\input{tables/llava}

\input{tables/qwen}

\subsection{Pruning on Quantized Models}

For VLMs, we can begin pruning image tokens at two stages: one at the vision encoder and the other at the language model component. We input the image and text pairs to collect each output token's attention score distribution $\alpha$ in different layers. For the vision encoder, we use $\alpha_{\mathrm{img}}^{i,j}$ to donate the attention score of the $i$-th token in the $j$-th layer. For the language model part, we follow FastV~\cite{chen2024image} to use $\alpha_{\mathrm{sys}}^{i,j}$, $\alpha_{\mathrm{txt}}^{i,j}$, and $\alpha_{\mathrm{out}}^{i,j}$ for system prompt, image tokens, user instruction, and output tokens. For the vision encoder we have:

\begin{equation}
\alpha_{\mathrm{img}}^{i,j} = 1,
\end{equation}
For the language model we have:
\begin{equation}
\alpha_{\mathrm{sys}}^{i,j}
+ \alpha_{\mathrm{img}}^{i,j}
+ \alpha_{\mathrm{ins}}^{i,j}
+ \alpha_{\mathrm{out}}^{i,j}
= 1 
\end{equation}

Then we calculate the token attention score to obtain the salient tokens, take image token for example, the attention score of image tokens in layer~$j$ is:

\begin{equation}
\lambda_{\mathrm{img}}^j
= \frac{\sum_{i=1}^n \alpha_{\mathrm{img}}^{i,j}}{N_\mathrm{img}},
\end{equation}
where $N_\mathrm{img}$ is the number of image tokens, $n$ is the number of tokens in the response.







%% file: tables/llama.tex
\begin{table*}[h!]
\centering
\vspace{-3mm}
\resizebox{1.0\textwidth}{!}{
\begin{tabular}{lcccccccc}
\toprule
\textbf{Benchmark} & \textbf{FP} & \textbf{Vis}& \textbf{Adp} & \textbf{Lm}& \textbf{Vis+Adp} & \textbf{Vis+Lm} & \textbf{Lm+Adp}& \textbf{Vis+Adp+Lm}   \\
\midrule
MME Perception & 1446.81  & 1094.94 & 1446.47 & 1097.67 & 1155.65 & 653.20 &  1077.34 & 781.51 \\ 
\midrule
MME Cognition & 341.42  & 301.78 & 334.28 & 248.21 & 282.50 & 149.28 & 155.00 &  222.85 \\
\midrule
ScienceQA-IMG & 85.82  & 71.44 & 85.87 & 12.99 & 70.50 & 7.34 & 11.35 & 17.40 \\ 
\midrule
MMMU & 42.78  & 34.89 & 42.78 & 25.56 & 33.89 & 26.89 & 28.56 & 24.56 \\ 
\midrule
VizWiz-VQA & 59.72 & 55.31 & 59.81 & 35.34 & 55.73 & 40.84 & 51.11 & 22.93 \\ 
\bottomrule 
\end{tabular}
}
\vspace{-2mm}
\caption{\textbf{Quantization on different components of Llama 3.2-Vision instruction 11B with weight 1 to 1.1 bit}. The vision model exhibits high sensitivity to quantization; the adaptor/projector exhibits less sensitivity to quantization, barely affecting the performance; the language model exhibits considerable sensitivity to quantization. FP: Full precision. Vis: Vision encoder. Adp: Adapt layer. Lm: Language model. }
\label{table:components_llama}
\vspace{-2mm}
\end{table*}

%% file: tables/llava.tex
\begin{table*}[h!]
\centering
\resizebox{1.0\textwidth}{!}{
\begin{tabular}{lcccccccc}
\toprule
\textbf{Benchmark} & \textbf{FP} & \textbf{Vis}& \textbf{Adp} & \textbf{Lm}& \textbf{Vis+Adp} & \textbf{Vis+Lm} & \textbf{Lm+Adp}& \textbf{Vis+Adp+Lm}   \\
\midrule
MME Perception & 1578.39 & 1180.35 & 1538.72 & 1024.15 & 1140.72 & 793.58 & 1043.01 & 813.89 \\ 
\midrule
MME Cognition & 418.21 & 304.64 & 392.5 & 150.36 & 287.14 & 163.57 & 146.43 & 178.57  \\
\midrule
ScienceQA-IMG & 95.84 & 83.04 & 95.79 & 72.83 & 82.94  & 61.28 &  74.71 & 63.81 \\ 
\midrule
MMMU & 49.56 & 44.33 & 49.00 & 31.33 & 44.33 & 29.67 & 30.67 & 28.78 \\ 
\midrule
VizWiz-VQA & 60.38 & 56.34 & 60.67 & 56.91 & 56.39 & 52.66 & 57.17 & 52.81 \\ 
\bottomrule 
\end{tabular}
}
\vspace{-2mm}
\caption{\textbf{Quantization on different components of Llava-One-Vision 7B with weight 1 to 1.1 bit}. Same conclusion as in Table~\ref{table:components_llama}. Full precision. Vis: Vision encoder. Adp: Adapt layer. Lm: Language model. }
\label{table:components_llava}
\vspace{-2mm}
\end{table*}

%% file: tables/qwen.tex
\begin{table*}[h!]
\centering
\resizebox{1.0\textwidth}{!}{
\begin{tabular}{lcccccccc}
\toprule
\textbf{Benchmark} & \textbf{FP} & \textbf{Vis}& \textbf{Adp} & \textbf{Lm}& \textbf{Vis+Adp} & \textbf{Vis+Lm} & \textbf{Lm+Adp}& \textbf{Vis+Adp+Lm}   \\
\midrule
MME Perception & 1683.88 & 1232.54 & 1707.72 & 1204.31 & 1226.47 & 838.91 & 1222.28 & 828.54 \\ 
\midrule
MME Cognition & 653.21 & 343.21 & 660.71 & 341.07 & 330.36 & 208.21 & 344.64 &  195.36 \\
\midrule
ScienceQA-IMG & 77.29 & 71.44 & 76.90 & 62.62 & 71.00 & 58.85 & 62.77 & 59.49 \\ 
\midrule
MMMU & 51.00 & 43.78 & 50.78 & 35.33 & 42.67  & 31.00 & 35.33  & 30.89 \\ 
\midrule
VizWiz-VQA & 70.43 & 56.10 & 70.30 & 59.85 & 55.99 & 52.75 & 59.72 & 52.72 \\ 
\bottomrule 
\end{tabular}
}
\vspace{-2mm}
\caption{\textbf{Quantization on different components of Qwen2.5-VL-7B-Instruct with weight 1 to 1.1 bit}. Same conclusion as in Table~\ref{table:components_llama}. FP: Full precision. Vis: Vision encoder. Adp: Adapt layer. Lm: Language model. }
\vspace{-2mm}
\label{table:sota_qwen}
\end{table*}

%% file: sec/4_experiment.tex
\section{Results and Comparisons}

\subsection{Baseline Model and Datasets}

\input{tables/llama_sota}
\input{tables/llava_sota}

\input{tables/qwen_sota}

To demonstrate the effectiveness of our proposed method, we compare our algorithm with the state-of-the-art (SOTA) methods on 4 dataset benchmarks and 3 models. Datasets include: MME~\cite{fu2024mmecomprehensiveevaluationbenchmark} that focuses on perception and cognition abilities; MMMU~\cite{yue2023mmmu} that focuses on college-level subject knowledge and deliberate reasoning, Science Question Answering (ScienceQA)~\cite{lu2022learn} that focuses on science topics, and VizWiz-VQA~\cite{gurari2018vizwiz} that on predicting the answer to a visual question and predict whether a visual question cannot be answered. Models include Llama 3.2-Vision instruction 11B~\cite{llama32vision}, Llava-One-Vision 7B ~\cite{liu2024llavanext, li2024llava}, and Qwen2.5-VL-7B-Instruct ~\cite{Qwen2.5-VL}. For all the experiments, we use 64 samples for calibration. We use 2 bits for salient weights and 1 bit for non-salient weights. Salient weights account up to 5\% of the total weights.
Ablation study, quantized weights storage and bitwidth calculation, statistical analysis, saliency threshold determination, please refer to Supplementary.

\subsection{Quantization on Different Components}

As shown in Table~\ref{table:components_llama}~\ref{table:components_llava}~\ref{table:components_qwen}, when quantizing only a single component, the vision model exhibits high sensitivity to quantization (on average); the adaptor/projector exhibits less sensitivity to quantization, barely affecting the performance; the language model exhibits considerable sensitivity to quantization.

\subsection{Comparison with SOTA Methods}



For Llama 3.2-Vision instruction 11B~\cite{llama32vision}, Table~\ref{table:sota_llama} summarizes the quantization performance of our Bi‐VLM method (under both language‐only and whole‐model quantization) compared to the state‐of‐the‐art AWQ and BiLLM approaches. Notably, 
our Bi-VLM outperforms SOTA by 4\%-47\% on all settings. Although Bi‐VLM shows somewhat lower performance on MME Cognition compared with BiLLM, the overall results confirm that Bi‐VLM consistently preserves most of the full‐precision accuracy across diverse benchmarks—frequently improving upon competing low‐bit methods—while substantially reducing model size and computational overhead.


For Llava-One-Vision 7B ~\cite{liu2024llavanext, li2024llava}, as shown in Table~\ref{table:sota_llama}, 
our Bi-VLM outperforms SOTA by 3\%-20\% on all settings, these results underscore that our Bi‐VLM quantization strategy substantially narrows the gap to full‐precision performance—often beating previous low‐bit methods by a large margin—across diverse vision‐language tasks.

For Qwen2.5-VL-7B-Instruct ~\cite{Qwen2.5-VL}, as shown in Table~\ref{table:sota_qwen}, 
our Bi-VLM outperforms SOTA by 4.5\%-12\% on all settings. Overall, these results demonstrate that our Bi‐VLM quantization strategy consistently outperforms prior low‐bit methods across diverse vision‐language tasks based on aggressive bitwidth reduction.


\subsection{Pruning on Quantized Model}

\begin{figure}[htb]
    \centering
    \includegraphics[width=1.\linewidth]{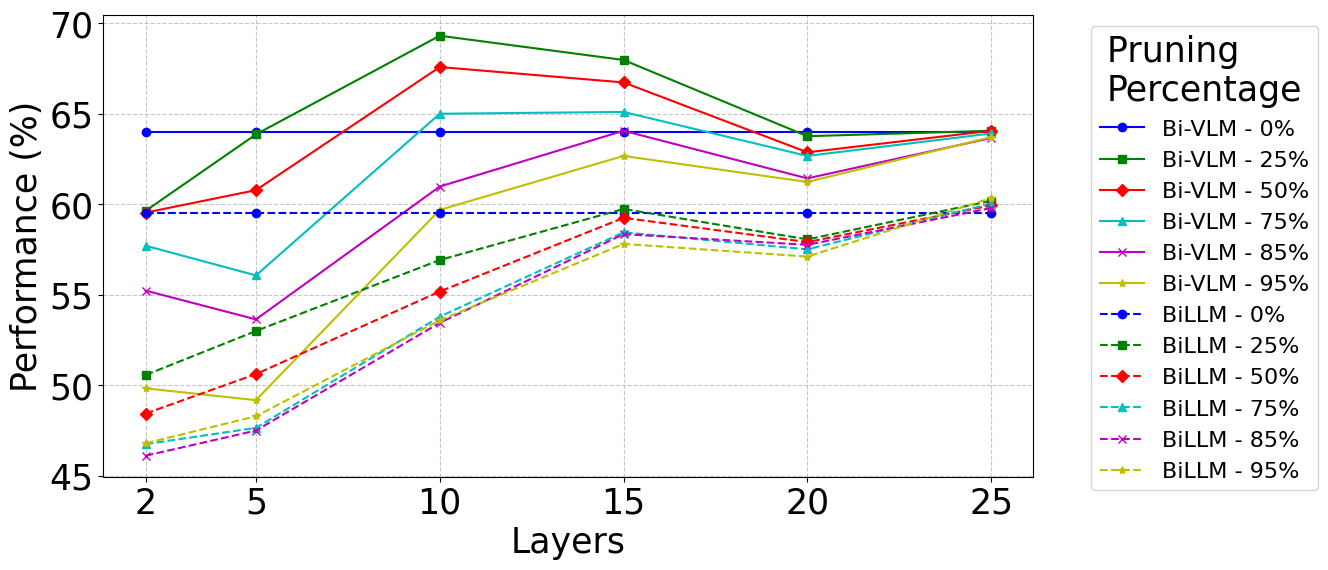}  
    \vspace{-4mm}
    \caption{Pruning on Bi-VLM quantized model and BiLLM quantized model. After layer 10, we can observe that there is redundancy of image tokens around 95\% in the quantized models. Our Bi-VLM exhibits better performance.}
    \label{fig:fig_prune}
    \vspace{-4mm}
\end{figure}

We also perform image token pruning on the quantized VLM (take Qwen2.5-VL-7B-Instruct for example), as shown in Figure~\ref{fig:fig_prune}, more pruning results please refer to Supplementary \ref{appendix:C}. Language tokens are more critical to model accuracy, as pruning them causes sharper declines in performance. Pruning image tokens in the language model is preferable to pruning in the vision encoder, since textual context can more effectively guide which image tokens to retain. Finally, the quantized model indicates a high degree of image token redundancy (90–99\%), allowing for aggressive compression with minimal performance impact.

%% file: tables/llama_sota.tex
\begin{table*}[h!]
\centering
\resizebox{1.0\textwidth}{!}{
\begin{tabular}{lccccccc}
\toprule
\textbf{Benchmark} & \textbf{FP} & \textbf{AWQ-L}& \textbf{BiLLM-L} & \textbf{Bi-VLM-L (Ours)}& \textbf{AWQ-all} & \textbf{BiLLM-all} & \textbf{Bi-VLM-all (Ours)} \\
\midrule
MME Perception & 1446.81  & 0.00 & 1096.67 & 1315.84 (\textbf{219.17$\uparrow$})  &  - &  781.51 &  1308.40 (\textbf{526.89$\uparrow$})  \\ 
\midrule
MME Cognition & 341.42  & 0.00 & 248.21 & 171.43 (\textbf{76.78$\downarrow$})  &  - & 222.85  &  196.07 (\textbf{26.78$\downarrow$}) \\
\midrule
ScienceQA-IMG & 85.82 & 0.00 & 11.01 & 58.75 (\textbf{47.74$\uparrow$})   & -  &  17.40 &  58.35 (\textbf{45.95$\uparrow$}) \\ 
\midrule
MMMU & 42.78  & 24.33 & 25.56 & 36.42 (\textbf{10.86$\uparrow$})  &  - &  24.56 &  33.27 (\textbf{8.71$\uparrow$})  \\ 
\midrule
VizWiz-VQA & 59.72  & 0.00 & 35.36 & 39.33 (\textbf{3.97$\uparrow$})  & -  & 22.93  &   47.18 (\textbf{24.25$\uparrow$})  \\ 
\bottomrule 
\end{tabular}
}
\vspace{-2mm}
\caption{\textbf{SOTA comparison on Llama 3.2-Vision instruction 11B with weight 1 to 1.1 bit}. For the language model part, our Bi-VLM outperforms the SOTA by 4\%-47\%. For the overall VLM,  our Bi-VLM outperforms the SOTA by 8\%-45\%. FP: Full precision. L: Language model, all: the whole VLM model.}
\label{table:sota_llama}
\vspace{-4mm}
\end{table*}


%% file: tables/llava_sota.tex
\begin{table*}[h!]
\centering
\resizebox{1.0\textwidth}{!}{
\begin{tabular}{lcccccccc}
\toprule
\textbf{Benchmark} & \textbf{FP} & \textbf{AWQ-L}& \textbf{BiLLM-L} & \textbf{Bi-VLM-L (Ours)}& \textbf{AWQ-all} & \textbf{BiLLM-all} & \textbf{Bi-VLM-all (Ours)} \\
\midrule
MME Perception & 1578.39 & 0.00 & 1024.15 & 1457.68 (\textbf{433.53$\uparrow$}) & -  & 813.89 & 1063.07 (\textbf{249.18$\uparrow$})\\ 
\midrule
MME Cognition & 418.21 & 0.00 & 150.36 & 340.71 (\textbf{190.35$\uparrow$}) & -  & 178.57 & 272.21 (\textbf{93.64$\uparrow$})\\
\midrule
ScienceQA-IMG & 95.84 & 0.00 & 72.83 & 93.55 (\textbf{20.72$\uparrow$}) & -   & 63.81 & 83.43 (\textbf{19.62$\uparrow$})\\ 
\midrule
MMMU & 49.56 & 25.33 & 31.11 & 44.33 (\textbf{13.22$\uparrow$}) & - & 28.78 & 39.12 (\textbf{10.34$\uparrow$})\\ 
\midrule
VizWiz-VQA & 60.38 & 0.00 & 56.91 & 60.10 (\textbf{3.19$\uparrow$}) & - & 52.81 & 57.36 (\textbf{4.55$\uparrow$})\\ 
\bottomrule 
\end{tabular}
}
\caption{\textbf{SOTA comparison on Llava-One-Vision 7B with weight 1 to 1.1 bit}. For the language model part, our Bi-VLM outperforms the SOTA by 3\%-20\%. For the overall VLM,  our Bi-VLM outperforms the SOTA by 4\%-19\%. FP: Full precision. L: Language model, all: the whole VLM model.}
\label{table:sota_llava}
\end{table*}

%% file: tables/qwen_sota.tex
\begin{table*}[h!]
\centering
\resizebox{1.0\textwidth}{!}{
\begin{tabular}{lccccccc}
\toprule
\textbf{Benchmark} & \textbf{FP} & \textbf{AWQ-L}& \textbf{BiLLM-L} & \textbf{Bi-VLM-L (Ours)}& \textbf{AWQ-all} & \textbf{BiLLM-all} & \textbf{Bi-VLM-all (Ours)} \\
\midrule
MME Perception & 1683.88 & 0.00 & 1204.31 & 1690.67 (\textbf{486.36$\uparrow$}) & - & 828.54 & 1458.70 (\textbf{630.16$\uparrow$}) \\ 
\midrule
MME Cognition & 653.21 & 0.00 & 341.07 & 621.79 (\textbf{280.72$\uparrow$}) & - &  195.36 & 559.64 (\textbf{364.28$\uparrow$})\\
\midrule
ScienceQA-IMG & 77.29 & 0.00 & 62.62 & 68.32 (\textbf{5.70$\uparrow$}) & - & 59.49 & 64.01 (\textbf{4.52$\uparrow$})\\ 
\midrule
MMMU & 51.00 & 25.44 & 35.33 & 46.00 (\textbf{10.67$\uparrow$}) & -  & 30.89 & 42.89 (\textbf{12.00$\uparrow$})\\ 
\midrule
VizWiz-VQA & 70.43 & 0.00 & 59.85 & 66.20 (\textbf{6.35$\uparrow$}) & - & 52.72 & 62.96 (\textbf{10.24$\uparrow$}) \\ 
\bottomrule 
\end{tabular}
}
\caption{\textbf{SOTA comparison on Qwen2.5-VL-7B-Instruct with weight 1.1 bit}. For the language model part, our Bi-VLM outperforms the SOTA by 5\%-10\%. For the overall VLM,  our Bi-VLM outperforms the SOTA by 4\%-12\%. FP: Full precision. L: Language model, all: the whole VLM model.}
\label{table:components_qwen}
\vspace{-5mm}
\end{table*}

%% file: sec/5_conclusion.tex
\section{Conclusions, Limitations, and Future Work}
We present a novel ultra-low-bit quantization method for VLMs. Our approach successfully bridges the gap between the computational demands of vision-language models and the practical limitations of ultra-low-bit precision. 
Our results demonstrate that Bi-VLM outperforms the state-of-the-art methods on both language and overall vision-language models. 
Our approach has some limitations.  We did not explore our method on large bitwidth, like 4 bits or 8 bits. It may be useful to explore combinations of different bitwidths. We would like to evaluate the performance on more tasks and in hardware-constrained environments.

%% file: sec/6_appendix.tex
\newpage
\appendix

\begin{figure*}[!h]
    \centering
    \includegraphics[width=1.\textwidth]{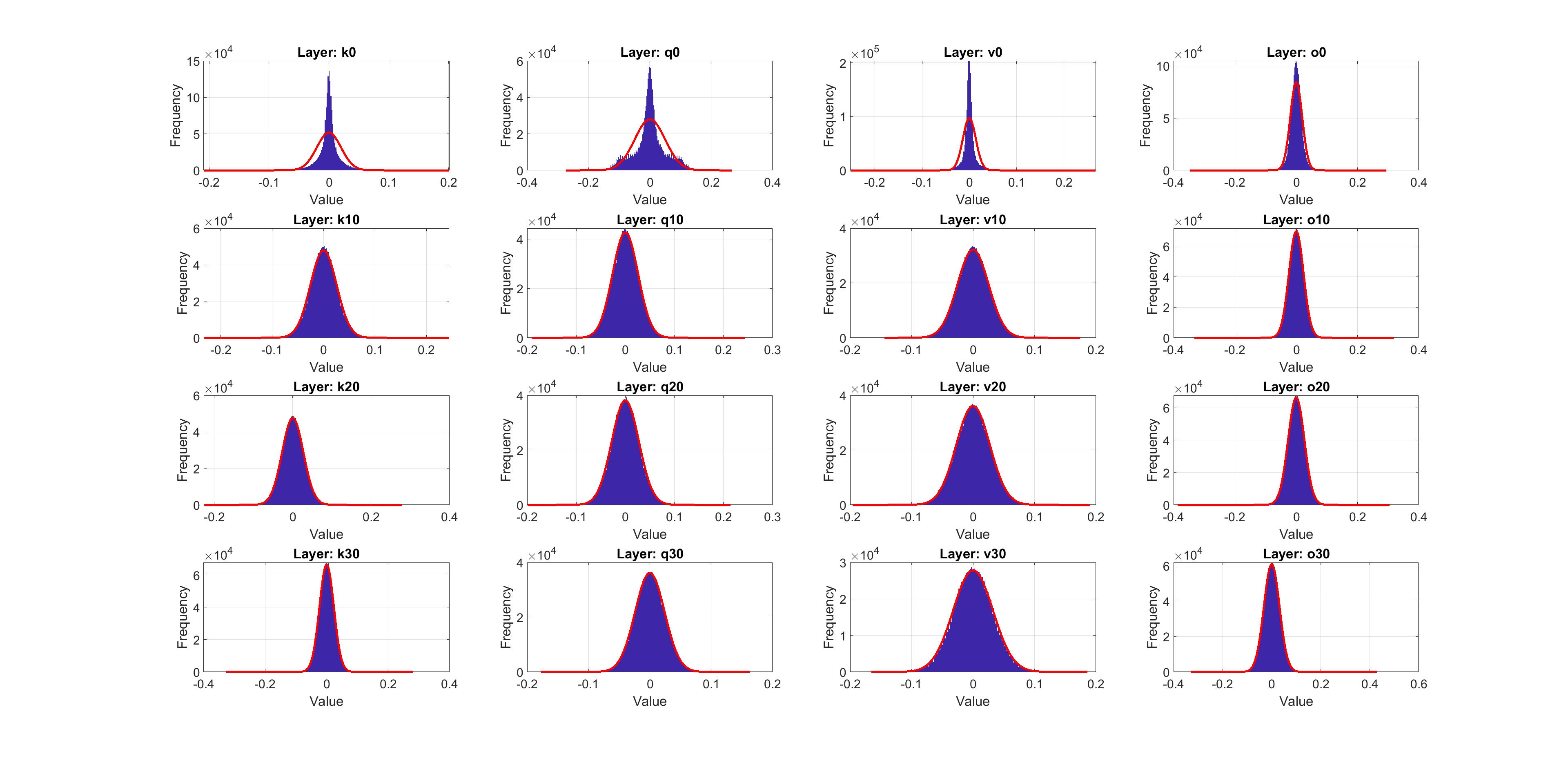}  
    \caption{Histogram Analysis and Fitted Gaussian Curve of Layers [0,10,20,30] of the Vision Model. The curve represents the fitted Gaussian distribution over the histogram bar plot. }
    \label{fig:fig1}
\end{figure*}
\section{Appendix}

\subsection{Memory-Efficient Quantization Pipeline}

Our quantization method partitions each model layer’s weight matrix into salient and unsalient weights, assigning higher bit precision to salient weights while binarizing the rest. By categorizing weights based on magnitude distributions, we enforce a strict scaling strategy that balances quantization accuracy and compression efficiency. To further optimize storage, we introduce codebook encoding to efficiently manage indexed weight representation, minimizing indexing overhead and ensuring the storage rate remains within the binarization scope.
\subsubsection{Quantized Weights Storage}

To evaluate the storage reduction of our quantization approach, we define compression threshold and select parameters \( N_{\text{uns}} \), \( N_b \), and \( p_l^{\text{sal}, \max} \) for each model layer. Compression efficiency is measured by the \textit{average storage bits per weight}. To achieve weight binarization goal, we set \( N_b = 2 \) for salient weight quantization and empirically choose \( p_l^{\text{sal}, \max} = 1\% \), assuming 16-bit storage per scaling factor, resulting in an average storage of 1.014 bits per weight.

Our method is fully parameterized, enabling trade-offs between accuracy and storage. The \textit{average storage bits per weight}, \( L_{\text{model}} \), is computed as:

\begin{equation}
    L_{\text{model}} = L_B + L_a
\end{equation}

where \( L_B \) represents binarized weight storage, and \( L_a \) accounts for scaling parameter overhead:

\begin{equation}
    L_B \leq 1 + (N_b - 1) p_l^{\text{sal}, \max} \quad \text{bits}
\end{equation}

\begin{equation}
    L_a = \frac{N_{\text{uns}} \cdot 16 + 16 \cdot m}{m \cdot n} \quad \text{bits}
\end{equation}

where \( m \) and \( n \) are the row and column dimensions of \( W_l \). The total storage requirement combines both components, ensuring efficient quantized representation.

\sloppy
\begin{figure*}[htb]
    \centering
    \includegraphics[width=1.0\textwidth]{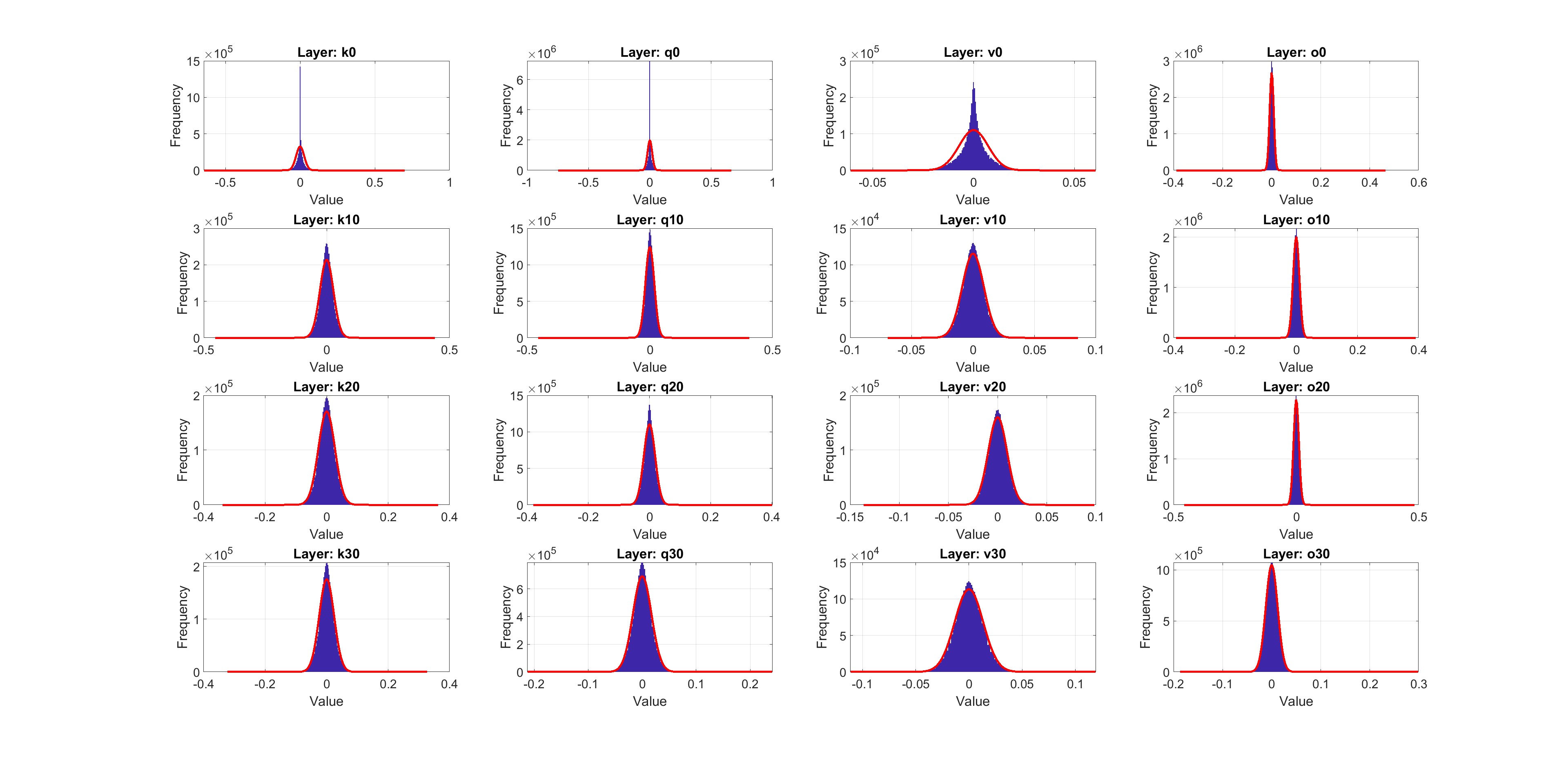}  
    \caption{Histogram Analysis and Fitted Gaussian Curve of Layers [0,10,20,30] of the Language Model. The curve represents the fitted Gaussian distribution over the histogram bar plot. }
    \label{fig:fig2}
\end{figure*}

\subsubsection{Codebook Encoding}

Upon completing model quantization, the next step is to efficiently store the quantized representations in memory. We propose an approach for encoding the indices of quantized weights based on their respective quantiles in the weight distribution. Given $N_{\text{uns}}$ unsalient weight groups and a single salient group, we aim to map weights to these $N_{\text{uns}}+1$ subsets while preserving the compression objective.

To optimize storage, we encode each weight group using a compressed bitwise format. Each weight is assigned a group code based on its quantile and stored in an index-encoded representation. Using bit-packing \cite{madhav2023bitpacking}, indices are compactly packed into a continuous data stream, improving storage efficiency and retrieval.

Defining \( L_{\text{i,max}} \) as the maximum number of bits allocated for encoding weight groups, we determine the maximum number of unsalient partitions while accounting for a single salient group:

\begin{equation}
N_{\text{uns,max}}(L_{\text{i,max}}) = 2^{L_{\text{i,max}}} - 3
\end{equation}

Following the bit-packing technique, the total number of bits, \( L_i \), required for storing the quantile-encoded indices of salient and non-salient weights, given a selected \( N_{\text{uns}} \), maximum salient percentile \( p_l^{\text{sal,max}} \), and unsalient proportion \( p_l^{\text{uns}} \), for layer \textit{l} is:

\begin{equation}
L_i = \left[ \sum_{\eta=1}^{L_{\text{i,max}}} \eta \cdot \min(2^\eta, N_{\text{uns}} - 2^\eta + 1) \right] p_l^{\text{uns}} + p_l^{\text{sal,max}} \cdot L_{\text{i,max}}
\end{equation}

This formulation allows dynamic adjustment of compression parameters through a defined threshold, translating into quantization parameters. Our bit-packing technique efficiently compresses index encoding, achieving an average binarized index encoding of 1.6 bits per index for \( N_{\text{uns}} = 5 \) partitions. And in practice, it can reach \( N_{\text{uns}} = 6 \) since you can leave one partition unmasked. 

It is important to note that index does not participate in the computation; actual calculations are executed solely with parameter weights. Therefore, additional hardware identification bits do not affect the acceleration effect of binary quantization.

\subsection{Statistical Analysis of Weights: Histograms and Gaussian Fit}
\label{appendix:A}

Understanding weight matrix statistics is essential for designing an efficient group-wise post-training quantization strategy. We analyze weight distributions through histograms, validating their tendency toward a Gaussian distribution. This supports our Gaussian assumption and highlights the effectiveness of a quantile-based approach for partitioning weights into consistent subsets. Since weight values vary significantly, saliency analysis is crucial for identifying critical weights. Our findings, observed across both Vision and Language Models, inform our post-training quantization approach. 

\paragraph {Key Observations} \

We analyze the element-wise weight histograms of both Vision and Language Models, using varying bin sizes based on layer dimensions. Our focus is on the key, query, and value projection layers in self-attention, as well as the final output layer. To approximate the weight distribution, we fit the closest Gaussian curve using the sample mean and variance. For both Vision and Language models, most weights exhibit a Gaussian-like distribution with a near-zero mean, consistent with prior findings on deep learning architectures and LLMs\cite{fang2020post,yu2020low,cholakov2023distributional}.  This suggests that weights remain close to their initial values, reinforcing the assumption that large-scale models preserve their initialized weight structure. Therefore, we adopt Gaussian assumption on layer weights distribution for both Vision and Language models.

As shown in Figures~\ref{fig:fig1} and~\ref{fig:fig2}, early self-attention layers exhibit sparser tails and significant deviations from the Gaussian trend, with weight values more concentrated around the mean compared to a standard normal distribution. To quantify this deviation, we conducted a secondary analysis using Kullback–Leibler divergence (KL) divergence to measure the discrepancy between the fitted Gaussian and the actual layer histograms across both Vision and Language Models.

\begin{figure}[htb]
    \centering
    \includegraphics[width=0.5\textwidth, height=0.3\textheight]{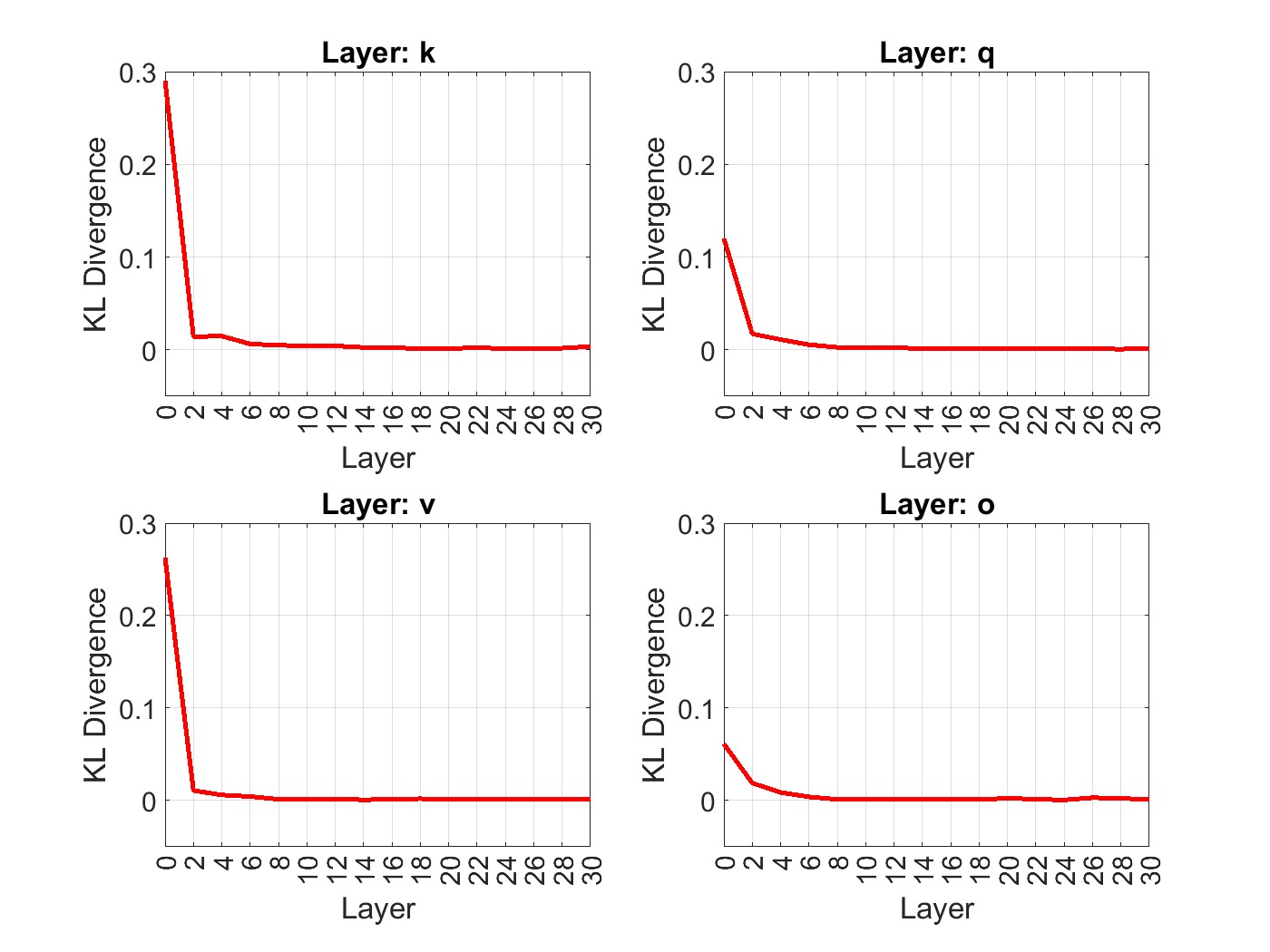}  
    \caption{KL divergence between weight histograms and fitted Gaussian distributions in a Vision Model. Early self-attention layers exhibit significant deviation from the Gaussian approximation compared to later layers.}
    \label{fig:fig3}
\end{figure}

\begin{figure}[htb]
    \centering
    \includegraphics[width=0.5\textwidth, height=0.3\textheight]{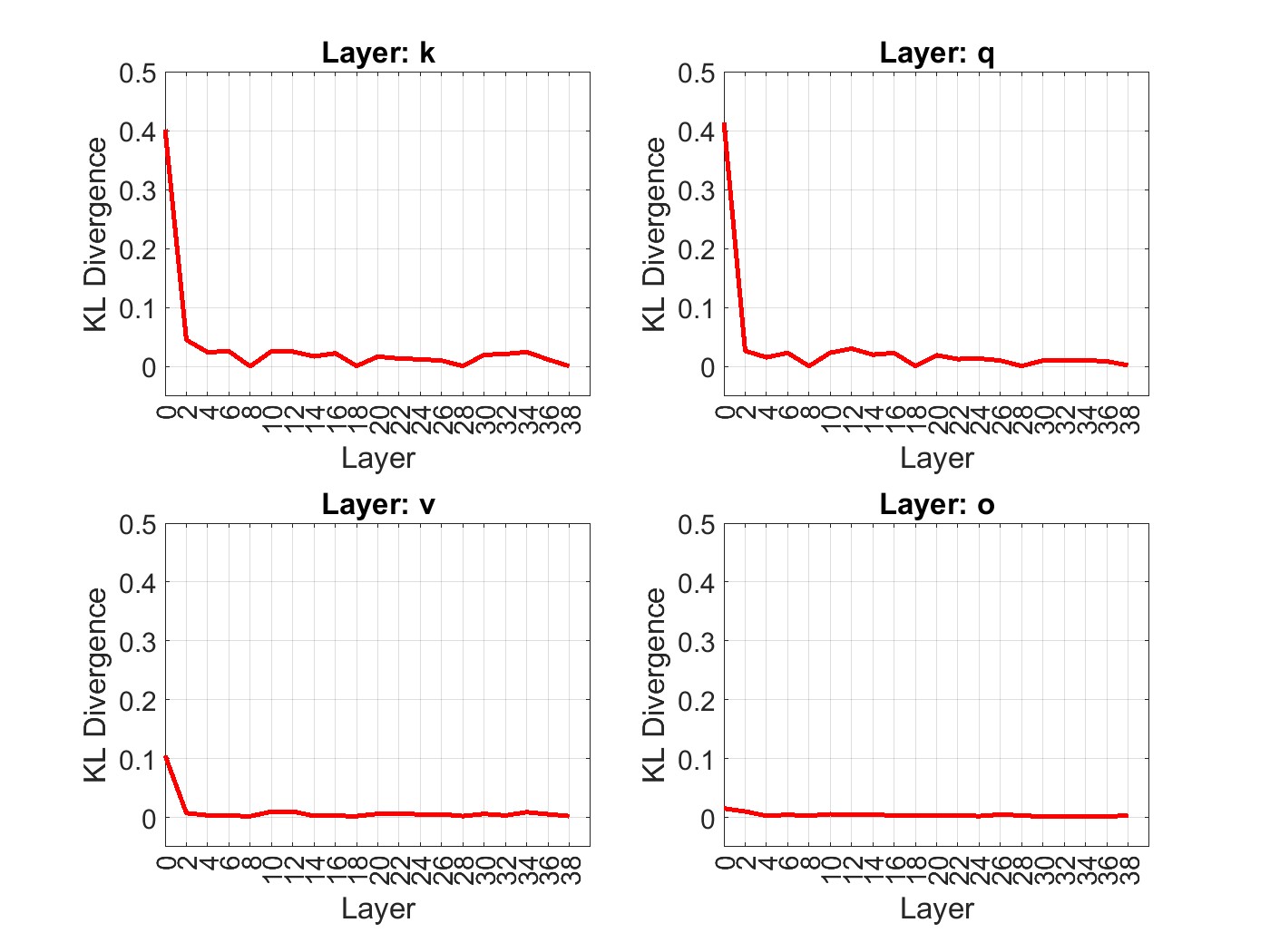}  
    \caption{KL divergence between weight histograms and fitted Gaussian distributions in a Language Model. Early self-attention layers exhibit significant deviation from the Gaussian approximation compared to later layers.}

    \label{fig:fig4}
\end{figure}

Our divergence analysis, depicted in Figures~\ref{fig:fig3} and~\ref{fig:fig4} indicate that only early self-attention layers deviate noticeably from the Gaussian distribution. However, by carefully tuning quantization parameters and employing multiple quantile-based divisions, this deviation can be effectively mitigated, ensuring robustness of our quantization strategy.
 \sloppy

\sloppy

Another key observation in weight distribution analysis is the impact of outliers. As seen in Figures~\ref{fig:fig1} and~\ref{fig:fig2}, the fitted distribution tails reveal a significant presence of outliers deviating from the symmetric Gaussian shape. The density of outliers varies across layers—some layers exhibit a high concentration, while others remain more stable. This insight motivated our dynamic outlier selection strategy, ensuring careful handling of these weights to improve quantization robustness.

\sloppy
\begin{figure}[htb]
    \centering
    \includegraphics[width=1.\linewidth]{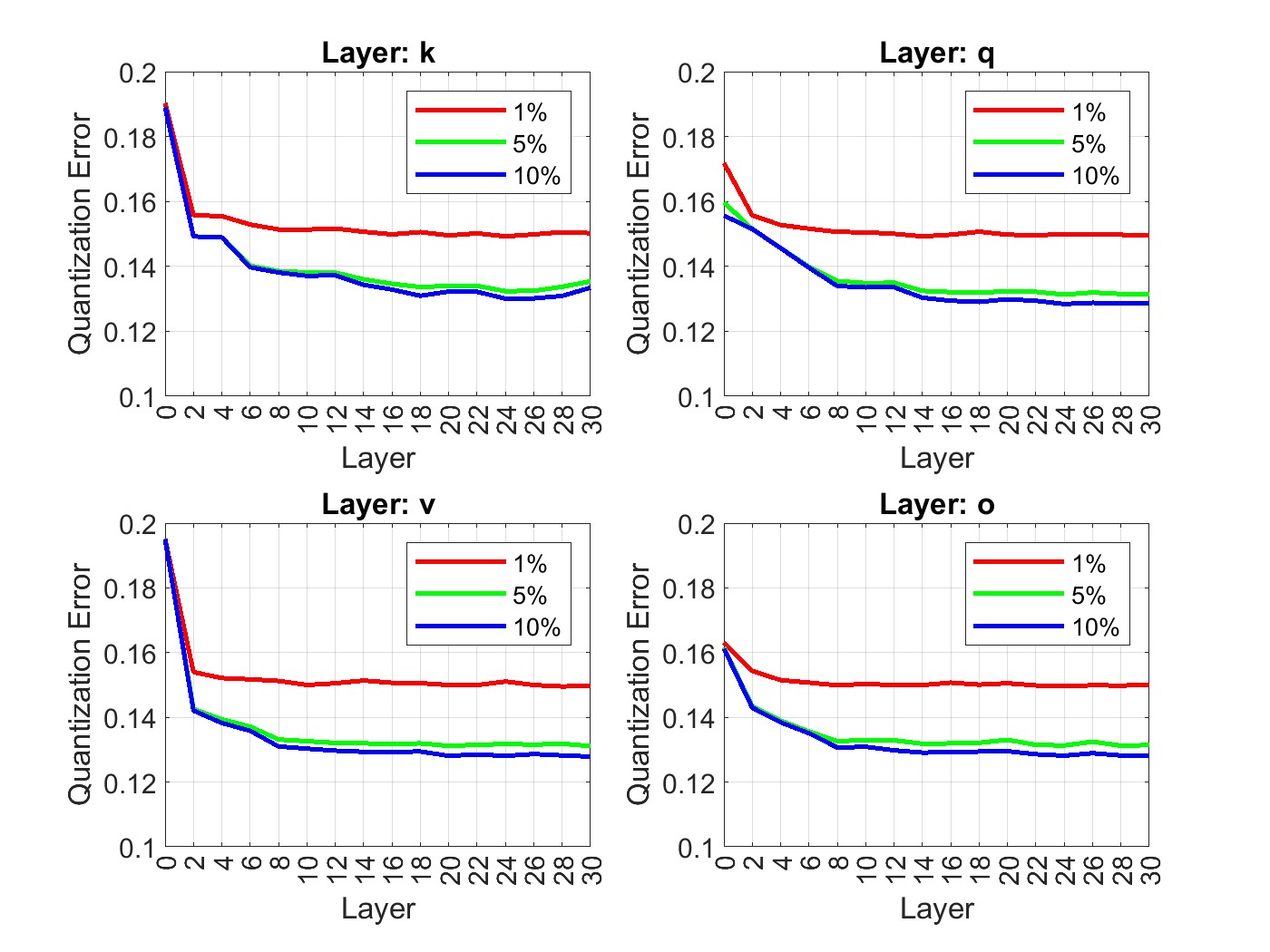}  
    \caption{ Bi-VLM Quantization Error Across Vision Model Layers for Varying Saliency Thresholds}

    \label{fig:fig5}
\end{figure}

\sloppy
\begin{figure}[htb]
    \centering
    \includegraphics[width=1.\linewidth]{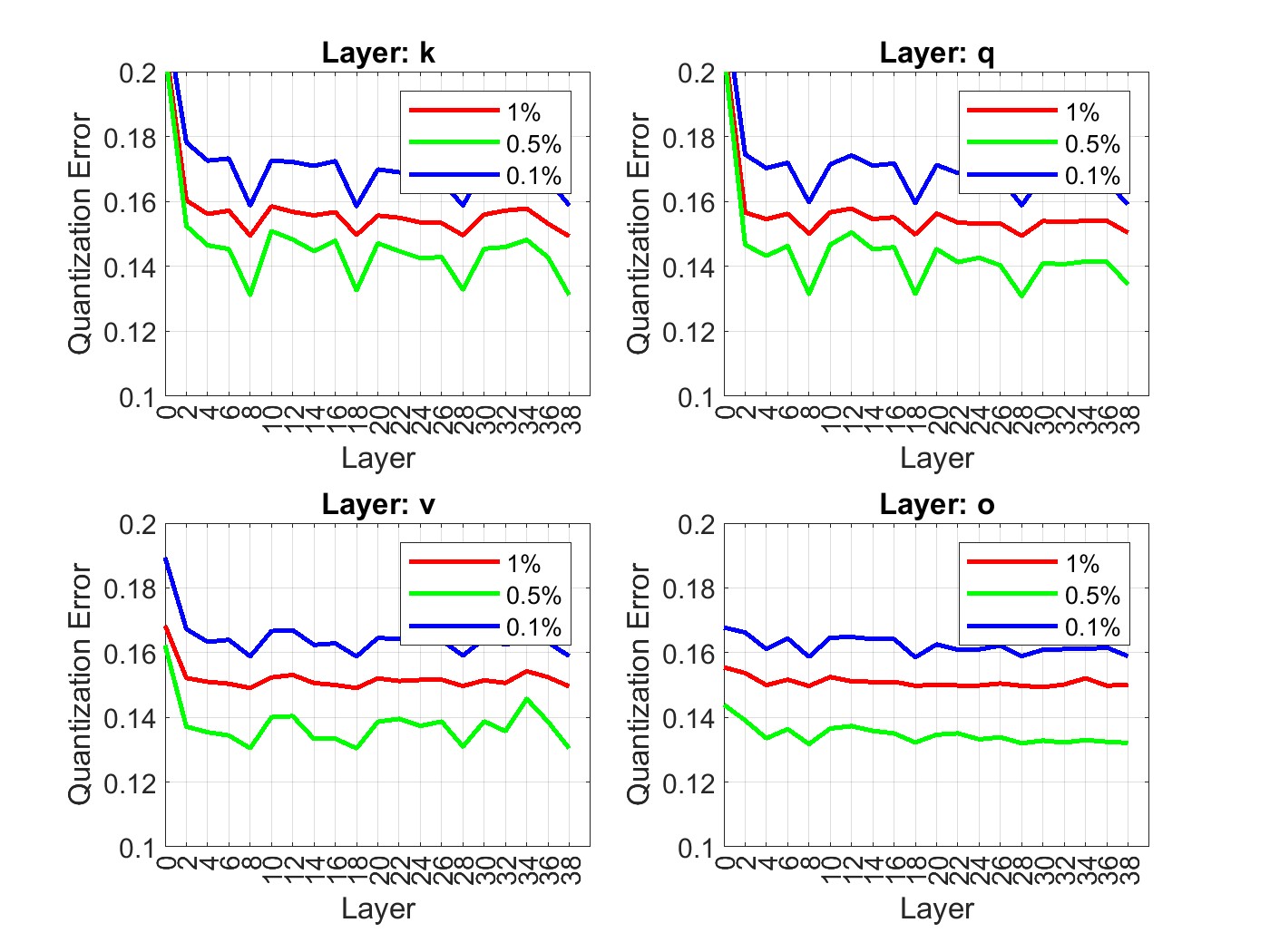}  
    \caption{ Bi-VLM Quantization Error Across Language Model Layers for Varying Saliency Thresholds}

    \label{fig:fig6}
\end{figure}

\subsection{Saliency Threshold Determination}
\label{appendix:B} 
Salient weights are defined based on their magnitude as outliers, requiring specialized handling and a distinct quantization strategy from unsalient weights. However, selecting too many salient weights can degrade model performance by affecting the scaling of unsalient weights, as salient weights are significantly larger in magnitude. To optimize this selection, our saliency search algorithm employs a bounded numerical optimization approach, where a threshold must be set to determine the optimal percentage of salient weights while minimizing quantization error. To refine this threshold, we conducted empirical analyses of our quantization algorithm under different maximum salient weight percentiles, allowing the search algorithm to explore various bounds. Additionally, since Vision Model weight matrices are smaller than those in Language Models, we assume that Vision Models can accommodate more salient weights while maintaining compression efficiency. Thus, our goal is to determine an optimal salient weight range that balances compression efficiency and quantization error across different models.

Figure~\ref{fig:fig5} presents the quantization performance under varying maximum salient percentile thresholds for the Vision Model. To meet the compression objective, we enforce binarization for at least 90\% of the weights, defining the salient weight range between 1\% and 10\%. At 1\% saliency, quantization error remains relatively high due to the limited number of weights assigned more than 1 bit. However, increasing the saliency threshold to 5\% significantly reduces quantization error, while further increasing it to 10\% offers only marginal improvements. To balance compression efficiency and quantization accuracy, we set the maximum saliency percentage for the Vision Model at 5\%.

Similarly, Figure~\ref{fig:fig6} shows the analysis for the Language Model, where we impose a stricter compression constraint—at least 99\% of the weights must be binarized. Consequently, the salient weight threshold is limited to 1\% at most. From quantization error analysis, we observe that handling approximately 1\% of weights as salient is necessary to maintain accuracy while meeting the compression and binarization objectives. Therefore, we select 1\% as the maximum saliency threshold for the Language Model.

\subsection{More results on Pruning in ScienceQA-IMG}
\label{appendix:C} 
\input{tables/pruning}
\textbf{Experimental Setting}

We conduct our experiments using the Llama-3.2-11B-Vision-Instruct, 
Llava-onevision-qwen2-7B-ov, and Qwen2.5-VL-7B-Instruct models, evaluating  pruning strategies under various quantization methods. Specifically, we employ Bi-VLM (ours) and BiLLM quantization techniques, which enforce binarization constraints at different saliency thresholds for both vision and language components. 

We evaluated the performance of the pruned models on ScienceQA-IMG, a dataset designed for multi-modal reasoning that requires both vision and language understanding. This dataset ensures a robust assessment of the model's ability to retain comprehension capabilities after pruning.

\noindent \textbf{Pruning Strategies}

We explore three pruning strategies in our experiments. The first approach involves only pruning the vision model while keeping the language model intact. The second strategy focuses on pruning only the image tokens in the language model, leaving the vision model unpruned. Lastly, we implement joint pruning, where both the image tokens in the vision model and the text tokens in the language model are pruned. Additionally, we evaluate layer-wise pruning with different ratios, ranging from 25\% to 99\%, to analyze the trade-offs between compression and accuracy. 

\noindent \textbf{Evaluation Metrics}

Model performance is assessed using three key metrics: accuracy, which measures task performance, pruning ratio, which represents the proportion of removed weights, and pruning layer, which indicates where pruning begins in the model.

\noindent \textbf{Results}

We observed that as pruning is applied to deeper layers, the model's accuracy becomes less sensitive to the pruning ratio, suggesting that deeper layers may have more redundancy. From Tables ~\ref{table:9},~\ref{table:11}, ~\ref{table:13},~\ref{table:16},~\ref{table:18}, we found that the language is more sensitive for pruning, the accuracy drops sharply as pruning increases. This indicates that pruning the text token has a much more severe impact on model performance than pruning image tokens. From pairs (Table~\ref{table:7}, Table~\ref{table:7}), (Table~\ref{table:14}, Table~\ref{table:15}), we found that pruning the image token in the language model is better than pruning in the vision encoder. This may be because the text information helps select the important image tokens, which enable a few important image tokens to be enough. From Table ~\ref{table:10} and ~\ref{table:12}, we found that the quantized model has 90\%-99\% image token redundancy. 

Overall, these findings highlight that language tokens are far more critical to model performance than image tokens, as pruning text tokens dramatically reduces accuracy. In contrast, pruning image tokens does not cause as severe a performance drop; in fact, removing image tokens in the language model may be preferable to pruning them in the vision encoder, given that text information can guide which image tokens are most essential. Additionally, the quantized model analysis suggests that there is substantial redundancy in image tokens (up to 90–99\%), indicating that a high degree of compression is possible without substantially harming performance.

%% file: tables/pruning.tex
\begin{table}[h!]
\centering
\resizebox{1.0\columnwidth}{!}{
\begin{tabular}{lcccccccc}
\toprule
\textbf{Layers} & \textbf{25\%} & \textbf{50\%}& \textbf{75\%} & \textbf{86.32\%}& \textbf{95.02\%} & \textbf{99\%}  \\
\midrule
2 & 85.72  & 85.87 & 85.72 & 85.47 & 78.33 & 57.76 \\ 
\midrule
5 & 86.12  & 85.67 & 85.97 & 85.32 & 80.86 & 53.79 \\
\midrule
10 & 86.17  & 86.22 & 85.97 & 85.87 & 82.55 & 58.70  \\ 
\midrule
15 & 85.87  & 85.92 & 85.77 & 85.72 &  82.00 & 65.00  \\ 
\midrule
20 & 85.67  & 85.97 & 85.82 & 85.23 & 77.44 & / \\ 
\midrule
25 & 85.92 & 85.82 & 85.08 & 80.32 & 71.19 & 60.59 \\
\midrule
30 & 85.92 & 85.57 & 83.69 & 79.67 & 72.29 & 59.20 \\
\bottomrule 
\end{tabular}
}
\caption{\textbf{Llama:} Only prune the vision model, keep others unpruned. [FP: Evaluated on Full precision model]. The baseline accuracy for the full precision model is 86.22\%.}
\label{table:7}
\vspace{-2mm}
\end{table}


\begin{table}[h!]
\centering
\resizebox{1.0\columnwidth}{!}{
\begin{tabular}{lcccccccc}
\toprule
\textbf{Layers} & \textbf{25\%} & \textbf{50\%}& \textbf{75\%} & \textbf{86.32\%}& \textbf{95.02\%} & \textbf{99\%}  \\
\midrule
2 (3) & 85.87 & 85.97 & 85.62 & 85.72 & 82.99 & 79.42 \\
\midrule
5(8) & 85.87 & 85.87 & 85.52 & 84.73 & 83.39 & 83.39  \\
\midrule
10 (13) & 86.07 & 85.97 & 85.87 & 86.22 & 85.67 & 85.57   \\ 
\midrule
15 (18) & 85.82 & 86.07 & 86.02 & 86.02 & 86.17 & 86.27  \\ 
\midrule
20 (23) & 86.07 & 86.07 & 86.17 & 85.97 & 86.17 & 86.51 \\ 
\midrule
25 (28) & 86.07 & 85.97 & 86.02 & 86.07 & 86.27 & 86.37 \\
\midrule
30 (33) & 85.97 & 86.12 & 86.02 & 86.07 & 86.02 & 85.97 \\
\bottomrule 
\end{tabular}
}
\caption{\textbf{Llama:}  Keep the vision model part unpruned, and selective prune of only image tokens in the language model part [FP: Evaluated on Full precision model]. The number in parentheses denotes the cross-attention layers. Baseline accuracy for the full precision model is 86.22\%. Results indicate that pruning up to 86.32\% of the image tokens across language layers maintains performance above 84\%, suggesting significant redundancy in image tokens. However, extreme pruning levels (e.g., 95.02\%, 99\%) lead to substantial accuracy drops, highlighting the importance of retaining a minimal number of tokens to ensure effective model performance. 
}
\label{table:8}
\vspace{-2mm}
\end{table}


\begin{table}[h!]
\centering
\resizebox{1.0\columnwidth}{!}{
\begin{tabular}{lcccccccc}
\toprule
\textbf{Layers} & \textbf{25\%} & \textbf{50\%}& \textbf{75\%} & \textbf{86.32\%}& \textbf{95.02\%} & \textbf{99\%}  \\
\midrule
2 & 52.11 & 27.42 & 7.39 & 1.29 & 0.00 & 0.00  \\
\midrule
5 & 66.73 & 42.14 & 10.01 & 1.59 & 0.00 & 0.00  \\
\midrule
10 & 67.72 & 45.02 & 32.28 & 9.87 & 0.59 & 0.00   \\ 
\midrule
15 & 83.39 & 73.92 & 44.03 & 23.50 & 3.17 & 0.00  \\ 
\midrule
20 & 85.87 & 84.09 & 63.51 & 35.00 & 5.50 & 0.00 \\ 
\midrule
25 & 85.87 & 86.42 & 79.67 & 56.92 & 15.12 & 0.00 \\
\midrule
30 & 85.67 & 85.77 & 84.28 & 65.54 & 16.66 & 0.00 \\
\bottomrule 
\end{tabular}
}
\caption{\textbf{Llama:} Pruning on both the vision and language part (vision token in vision model, text token in language model) [FP: Evaluated on Full precision model]. The baseline accuracy for the full precision model is 86.22\%. Results show that pruning both image tokens in the vision encoder and text tokens in the language model significantly degrades performance compared to Table 8 and Table 7, where only image tokens in the language model were pruned. Specifically, accuracy drops sharply as pruning increases. This indicates that pruning the text token has a much more severe impact on model performance than pruning image tokens.}
\label{table:9}
\vspace{-2mm}
\end{table}


\begin{table}[h!]
\centering
\resizebox{1.0\columnwidth}{!}{
\begin{tabular}{lcccccccc}
\toprule
\textbf{Layers} & \textbf{24.98\%} & \textbf{49.97\%}& \textbf{74.96\%} & \textbf{86.32\%}& \textbf{95\%} & \textbf{99\%}  \\
\midrule
2 (3) & 21.42 & 21.52 & 21.37 & 21.62 & 21.67 & 20.08  \\
\midrule
5(8) & 21.91 & 21.42 & 22.06 & 20.87 & 20.92 & 14.13  \\
\midrule
10 (13) & 21.67 & 21.62 & 22.31 & 22.21 & 21.32 & 15.77   \\ 
\midrule
15 (18) & 21.37 & 21.22 & 21.81 & 21.72 & 21.42 & 21.67  \\ 
\midrule
20 (23) & 21.57 & 21.67 & 22.11 & 22.06 & 21.67 & 22.06  \\ 
\midrule
25 (28) & 21.52 & 21.81 & 21.72 & 21.57 & 21.02 & 21.37  \\
\midrule
30 (33) & 21.37 & 21.72 & 21.62 & 21.62 & 21.27 & 20.33  \\
\bottomrule 
\end{tabular}
}
\caption{\textbf{Llama (BiLLM):}  Keep the vision model part unpruned, and selective prune of only image tokens in the language model part [vlm: Evaluated on quantized vision language model using BiLLM]. The baseline accuracy for the quantized model using BiLLM is 21.42\%. From this table, the quantized model has 90\%-99\% image token redundancy.
}
\label{table:10}
\vspace{-2mm}
\end{table}


\begin{table}[h!]
\centering
\resizebox{1.0\columnwidth}{!}{
\begin{tabular}{lcccccccc}
\toprule
\textbf{Layers} & \textbf{25\%} & \textbf{50\%}& \textbf{75\%} & \textbf{86.32\%}& \textbf{95.02\%} & \textbf{99\%}  \\
\midrule
2 & 8.53 & 5.01 & 1.14 & 0.00 & 0.00 & 0.00  \\
\midrule
5 & 10.51 & 5.65 & 0.30 & 0.00 & 0.00 & 0.00  \\
\midrule
10 & 19.34 & 11.80 & 0.89 & 0.00 & 0.00 & 0.00   \\ 
\midrule
15 & 19.93 & 11.45 & 1.34 & 0.00 & 0.00 & 0.00  \\ 
\midrule
20 & 21.22 & 8.48 & 4.07 & 2.88 & 0.00 & 0.00  \\ 
\midrule
25 & 21.52 & 21.02 & 17.85 & 8.58 & 0.40 & 0.00  \\
\midrule
30 & 21.72 & 20.82 & 15.27 & 12.74 & 1.24 & 0.00  \\
\bottomrule 
\end{tabular}
}
\caption{\textbf{Llama (BiLLM):} Pruning on both the vision and language part (vision token in vision model, text token in language model) [vlm: Evaluated on quantized vision language model using BiLLM]. The baseline accuracy for the quantized model using BiLLM is 21.42\%.
}
\label{table:11}
\vspace{-2mm}
\end{table}


\begin{table}[h!]
\centering
\resizebox{1.0\columnwidth}{!}{
\begin{tabular}{lcccccccc}
\toprule
\textbf{Layers} & \textbf{24.98\%} & \textbf{49.97\%}& \textbf{74.96\%} & \textbf{86.32\%}& \textbf{95\%} & \textbf{99\%}  \\
\midrule
2 (3) & 58.50 & 58.70 & 58.45 & 57.96 & 56.92 & 54.04  \\
\midrule
5(8) & 58.55 & 58.40 & 58.30 & 57.16 & 55.32 & 53.10  \\
\midrule
10 (13) & 58.65 & 58.55 & 57.71 & 57.91 & 56.37 & 55.58   \\ 
\midrule
15 (18) & 58.60 & 58.45 & 58.11 & 57.16 & 54.59 & 53.64  \\ 
\midrule
20 (23) & 58.60 & 58.60 & 58.01 & 57.56 & 57.26 & 57.36  \\ 
\midrule
25 (28) & 58.55 & 58.60 & 58.30 & 58.11 & 55.83 & 57.11  \\
\midrule
30 (33) & 58.90 & 58.70 & 58.40 & 58.30 & 58.11 & 58.70  \\
\bottomrule 
\end{tabular}
}
\caption{\textbf{Llama (Bi-VLM, Ours):} Keep the vision model part unpruned, and selective prune of only image tokens in the language model part [vlm: Evaluated on quantized vision language model using Bi-VLM]. The baseline accuracy for the quantized model using Bi-VLM is 58.35\%. From this table, the quantized model has 86\%-95\% image token redundancy.
}
\label{table:12}
\vspace{-2mm}
\end{table}


\begin{table}[h!]
\centering
\resizebox{1.0\columnwidth}{!}{
\begin{tabular}{lcccccccc}
\toprule
\textbf{Layers} & \textbf{25\%} & \textbf{50\%}& \textbf{75\%} & \textbf{86.32\%}& \textbf{95.02\%} & \textbf{99\%}  \\
\midrule
2 & 22.56 & 6.59 & 0.50 & 0.00 & 0.00 & 0.00  \\
\midrule
5 & 10.56 & 1.93 & 0.45 & 0.00 & 0.00 & 0.00  \\
\midrule
10 & 32.28 & 12.94 & 2.18 & 0.00 & 0.00 & 0.00  \\ 
\midrule
15 & 50.52 & 19.78 & 4.16 & 1.29 & 0.10 & 0.00  \\ 
\midrule
20 & 42.39 & 21.02 & 6.20 & 1.69 & 0.10 & 0.00  \\ 
\midrule
25 & 54.44 & 44.62 & 11.9 & 4.36 & 0.64 & 0.00  \\
\midrule
30 & 52.65 & 43.58 & 17.4 & 8.08 & 0.99 & 0.00  \\
\bottomrule 
\end{tabular}
}
\caption{\textbf{Llama (Bi-VLM, Ours):} Pruning on both the vision and language part (vision token in vision model, text token in language model) [vlm: Evaluated on quantized vision language model using Bi-VLM]. The baseline accuracy for the quantized model using Bi-VLM is 58.35\%.
}
\label{table:13}
\vspace{-2mm}
\end{table}


\begin{table}[h!]
\centering
\resizebox{1.0\columnwidth}{!}{
\begin{tabular}{lcccccccc}
\toprule
\textbf{Layers} & \textbf{27.4\%} & \textbf{45.14\%}& \textbf{73.1\%} & \textbf{86.3\%}& \textbf{93.3\%} & \textbf{98.76\%}  \\
\midrule
2 & 84.28 & 78.98 & 73.97 & 73.48 & 73.08 & 73.33  \\
\midrule
5 & 94.20 & 92.12 & 84.32 & 78.93 & 75.41 & 73.72  \\
\midrule
10 & 95.14 & 93.46 & 86.37 & 79.42 & 74.02 & 73.32  \\ 
\midrule
15 & 95.19 & 93.90 & 87.06 & 79.23 & 75.16 & 73.28  \\ 
\midrule
20 & 95.29 & 93.21 & 86.02 & 79.72 & 75.06 & 73.43 \\ 
\midrule
25 & 95.24 & 94.45 & 88.6 & 82.55 & 75.56 & 74.02  \\
\bottomrule 
\end{tabular}
}
\caption{\textbf{Llava:} Only prune the vision model, keep others unpruned. [FP: Evaluated on Full precision model]. The baseline accuracy for the full precision model is 95.84\%.
}
\label{table:14}
\vspace{-2mm}
\end{table}


\begin{table}[h!]
\centering
\resizebox{1.0\columnwidth}{!}{
\begin{tabular}{lcccccccc}
\toprule
\textbf{Layers} & \textbf{27.4\%} & \textbf{45.14\%}& \textbf{73.1\%} & \textbf{86.3\%}& \textbf{93.3\%} & \textbf{98.76\%}  \\
\midrule
2 & 94.55 & 93.55 & 88.80 & 83.19 & 78.93 & 75.61  \\
\midrule
5 & 94.60 & 92.91 & 90.68 & 88.60 & 84.78 & 76.65  \\
\midrule
10 & 96.03 & 95.84 & 92.02 & 87.95 & 83.69 & 78.43  \\ 
\midrule
15 & 95.79 & 95.69 & 95.29 & 94.70 & 92.56 & 87.75  \\ 
\midrule
20 & 95.93 & 95.93 & 95.93 & 95.88 & 95.93 & 95.93  \\ 
\midrule
25 & 95.88 & 95.88 & 95.88 & 95.88 & 95.88 & 95.88  \\
\bottomrule 
\end{tabular}
}
\caption{\textbf{Llava:}  Keep the vision model part unpruned, and selective prune of only image tokens in the language model part [FP: Evaluated on Full precision model]. The baseline accuracy for the full precision model is 95.84\%.
}
\label{table:15}
\vspace{-2mm}
\end{table}


\begin{table}[h!]
\centering
\resizebox{1.0\columnwidth}{!}{
\begin{tabular}{lcccccccc}
\toprule
\textbf{Layers} & \textbf{27.4\%} & \textbf{45.14\%}& \textbf{73.1\%} & \textbf{86.3\%}& \textbf{93.3\%} & \textbf{98.76\%}  \\
\midrule
2 & 58.70 & 45.71 & 33.66 & 16.81 & 4.71 & 0.00  \\
\midrule
5 & 80.22 & 59.10 & 32.77 & 21.86 & 11.01 & 3.47  \\
\midrule
10 & 82.25 & 58.50 & 39.07 & 36.19 & 31.43 & 11.75  \\ 
\midrule
15 & 93.01 & 87.41 & 68.62 & 47.40 & 40.06 & 31.73  \\ 
\midrule
20 & 95.24 & 93.06 & 75.66 & 63.96 & 64.25 & 34.26  \\ 
\midrule
25 & 95.19 & 94.50 & 88.45 & 82.50 & 74.17 & 27.37  \\
\bottomrule 
\end{tabular}
}
\caption{\textbf{Llava:} Pruning on both the vision and language part (vision token in vision model, text token in language model) [FP: Evaluated on Full precision model]. The baseline accuracy for the full precision model is 95.84\%.
}
\label{table:16}
\vspace{-2mm}
\end{table}


\begin{table}[h!]
\centering
\resizebox{1.0\columnwidth}{!}{
\begin{tabular}{lcccccccc}
\toprule
\textbf{Layers} & \textbf{27.4\%} & \textbf{45.14\%}& \textbf{73.1\%} & \textbf{86.3\%}& \textbf{93.3\%} & \textbf{98.76\%}  \\
\midrule
2 & 61.82 & 61.18 & 59.35 & 58.80 & 57.96 & 55.73  \\
\midrule
5 & 63.31 & 63.46 & 61.87 & 59.54 & 57.96 & 57.11  \\
\midrule
10 & 63.61 & 63.46 & 62.57 & 59.84 & 57.81 & 58.45  \\ 
\midrule
15 & 62.96 & 62.67 & 62.96 & 62.32 & 62.37 & 61.58  \\ 
\midrule
20 & 63.51 & 63.51 & 63.61 & 63.66 & 63.66 & 64.01  \\ 
\midrule
25 & 63.71 & 63.71 & 63.76 & 63.71 & 63.86 & 63.76  \\
\bottomrule 
\end{tabular}
}
\caption{\textbf{Llava (BiLLM):}  Keep the vision model part unpruned, and selective prune of only image tokens in the language model part [vlm: Evaluated on quantized vision language model using BiLLM]. The baseline accuracy for the quantized model using BiLLM is 63.81\%.
}
\label{table:17}
\vspace{-2mm}
\end{table}


\begin{table}[h!]
\centering
\resizebox{1.0\columnwidth}{!}{
\begin{tabular}{lcccccccc}
\toprule
\textbf{Layers} & \textbf{27.4\%} & \textbf{45.14\%}& \textbf{73.1\%} & \textbf{86.3\%}& \textbf{93.3\%} & \textbf{98.76\%}  \\
\midrule
2 & 42.59 &  37.68 & 24.74 & 11.60 & 3.82 & 0.00  \\
\midrule
5 & 46.90 & 38.52 & 34.61 & 26.08 & 13.98 & 0.00 \\
\midrule
10 & 53.40 & 43.78 & 36.54 & 32.77 & 34.01 & 6.30  \\ 
\midrule
15 & 56.47 & 52.80 & 37.88 & 36.49 & 35.45 & 28.06  \\ 
\midrule
20 & 62.96 & 62.02 & 54.83 & 51.21 & 53.35 & 36.69  \\ 
\midrule
25 & 63.06 & 62.52 & 59.35 & 58.06 & 57.76 & 47.15  \\
\bottomrule 
\end{tabular}
}
\caption{\textbf{Llava (BiLLM):} Pruning on both the vision and language part (vision token in vision model, text token in language model) [vlm: Evaluated on quantized vision language model using BiLLM]. The baseline accuracy for the quantized model using BiLLM is 63.81\%.
}
\label{table:18}
\vspace{-2mm}
\end{table}




\begin{table}[h!]
\centering
\resizebox{1.0\columnwidth}{!}{
\begin{tabular}{lcccccccc}
\toprule
\textbf{Layers} & \textbf{25\%} & \textbf{50\%}& \textbf{75\%} & \textbf{85\%}& \textbf{95\%} & \textbf{99\%}  \\
\midrule
2 & 79.23 & 78.83 & 76.85 & 73.43 & 69.41 & 67.13 \\ 
\midrule
5 & 70.10 & 69.26 & 67.23 & 65.69 & 62.42 & 61.03   \\
\midrule
10 & 69.56 & 70.40 & 68.52 & 65.49 & 64.95 & 63.71  \\ 
\midrule
15 & 78.14 & 78.58 & 77.24 & 77.39 & 77.09 & 76.80  \\ 
\midrule
20 & 76.10 & 75.71 & 75.71 & 75.41 & 75.41 & 75.06  \\ 
\midrule
25 & 77.69 & 77.79 & 77.34 & 77.44 & 77.09 & 77.19  \\
\bottomrule 
\end{tabular}
}
\caption{\textbf{Qwen:} Keep the vision model part unpruned, and selective prune of only image tokens in the language model part [FP: Evaluated on Full precision model]. The baseline accuracy for the full precision model is 77.29\%.
}
\label{table:19}
\vspace{-2mm}
\end{table}




\begin{table}[h!]
\centering
\resizebox{1.0\columnwidth}{!}{
\begin{tabular}{lcccccccc}
\toprule
\textbf{Layers} & \textbf{25\%} & \textbf{50\%}& \textbf{75\%} & \textbf{85\%}& \textbf{95\%} & \textbf{99\%}  \\
\midrule
2 & 50.57 & 48.44 & 46.75 & 46.11 & 46.80 & 47.00  \\ 
\midrule
5 & 53.00 & 50.62 & 47.65 & 47.5 & 48.29 & 42.49   \\
\midrule
10 & 56.92 & 55.18 & 53.79 & 53.45 & 53.59 & 53.00   \\ 
\midrule
15 & 59.74 & 59.25 & 58.45 & 58.35 & 57.81 & 58.21  \\ 
\midrule
20 & 58.06 & 57.91 & 57.51 & 57.76 & 57.11 & 57.46   \\ 
\midrule
25 & 60.19 & 59.94 & 60.04 & 59.79 & 60.34 & 60.14  \\
\bottomrule 
\end{tabular}
}
\caption{\textbf{Qwen (BiLLM):} Keep the vision model part unpruned, and selective prune of only image tokens in the language model part [vlm: Evaluated on quantized vision language model using BiLLM]. The baseline accuracy for the quantized model using BiLLM is 59.49\%.
}
\label{table:20}
\vspace{-2mm}
\end{table}




\begin{table}[h!]
\centering
\resizebox{1.0\columnwidth}{!}{
\begin{tabular}{lcccccccc}
\toprule
\textbf{Layers} & \textbf{25\%} & \textbf{50\%}& \textbf{75\%} & \textbf{85\%}& \textbf{95\%} & \textbf{99\%}  \\
\midrule
2 & 59.64 & 59.54 & 57.71 & 55.23 & 49.83 & 45.86  \\ 
\midrule
5 & 63.86 & 60.78 & 56.07 & 53.64 & 49.18 & 42.74   \\
\midrule
10 & 69.31 & 67.58 & 65.00 & 60.98 & 59.69 & 60.78   \\ 
\midrule
15 & 67.97 & 66.73 & 65.10 & 64.06 & 62.67 & 62.77  \\ 
\midrule
20 & 63.76 & 62.87 & 62.67 & 61.43 & 61.23 & 60.39   \\ 
\midrule
25 & 64.06 & 64.06 & 63.91 & 63.66 & 63.71 & 63.81  \\
\bottomrule 
\end{tabular}
}
\caption{\textbf{Qwen (Bi-VLM Ours):} Keep the vision model part unpruned, and selective prune of only image tokens in the language model part [vlm: Evaluated on quantized vision language model using Bi-VLM]. The baseline accuracy for the quantized model using Bi-VLM is 68.32\%.
}
\label{table:21}
\vspace{-2mm}
\end{table}

